\def\year{2019}\relax
\documentclass[letterpaper]{article} 
\usepackage{aaai19}  
\usepackage{times}  
\usepackage{helvet}  
\usepackage{courier}  
\usepackage{url}  
\usepackage{graphicx}  
\usepackage{footnote}
\usepackage{amsmath}
\usepackage{amsfonts}
\usepackage{bbm}
\usepackage{multirow}
\usepackage{booktabs}
\usepackage{footnote}
\usepackage{mwe} 
\usepackage{colortbl} 
\usepackage{subcaption} 
\usepackage[shortlabels]{enumitem} 
\usepackage{bm} 
\usepackage{dblfloatfix} 

\twocolumn
\begin{document}
%
\title{Subtask Gated Networks for Non-Intrusive Load Monitoring}

\author{Changho Shin\textsuperscript{1}, Sunghwan Joo\textsuperscript{2}, Jaeryun Yim\textsuperscript{1}, Hyoseop Lee\textsuperscript{1}, Taesup Moon\textsuperscript{2}, Wonjong Rhee\textsuperscript{3}\\
\textsuperscript{1}Encored Technologies, Seoul, Korea\\
\textsuperscript{2}Department of Electrical and Computer Engineering, Sungkyunkwan University, Suwon, Korea\\
\textsuperscript{3}Department of Transdisciplinary Studies, Seoul National University, Seoul, Korea\\
\textsuperscript{1}\{chshin, yim, hslee\}@encoredtech.com, \textsuperscript{2}\{shjoo840, tsmoon\}@skku.edu, \textsuperscript{3}wrhee@snu.ac.kr
}

\maketitle

\begin{abstract}
Non-intrusive load monitoring (NILM), also known as energy disaggregation, is a blind source separation problem where a household’s aggregate electricity consumption is broken down into electricity usages of individual appliances. In this way, the cost and trouble of installing many measurement devices over numerous household appliances can be avoided, and only one device needs to be installed. The problem has been well-known since Hart's seminal paper in 1992, and recently significant performance improvements have been achieved by adopting deep networks. In this work, we focus on the idea that appliances have on/off states, and develop a deep network for further performance improvements. Specifically, we propose a subtask gated network that combines the main regression network with an on/off classification subtask network. Unlike typical multitask learning algorithms where multiple tasks simply share the network parameters to take advantage of the relevance among tasks, the subtask gated network multiply the main network's regression output with the subtask's classification probability. When standby-power is additionally learned, the proposed solution surpasses the state-of-the-art performance for most of the benchmark cases. The subtask gated network can be very effective for any problem that inherently has on/off states.
\end{abstract}

\section{Introduction}
\label{sec:intro}
Non-intrusive load monitoring (NILM) was first proposed by Hart in 1992, and it is the process of estimating the power consumptions of individual appliances by using the aggregated electricity consumption as the only input \cite{hart1992nonintrusive}. Because the power consumption of multiple appliances are added together to form the aggregated consumption, and because the goal of NILM is to disaggregate each appliance's power consumption from the aggregate power consumption measured with a single sensor, NILM is also called energy disaggregation. 
The disaggregated energy consumption information can be used to provide feedback to the consumers and affect their energy consumption behaviors, and for instance
Neenan et al. showed that 15\% energy saving can be achieved \cite{neenan2009residential}.
Furthermore, NILM can be used for detecting malfunctioning appliances, designing energy incentives, and managing demand-response \cite{froehlich2010disaggregated}.
As an exemplary use case, utility companies can identify which homes are running a particular appliance (e.g., air conditioner) during the peak hours and take proper actions to reduce the power usage.

\begin{figure}[t] 
  \begin{subfigure}[b]{0.5\linewidth}
    \centering
    \includegraphics[width=\linewidth]{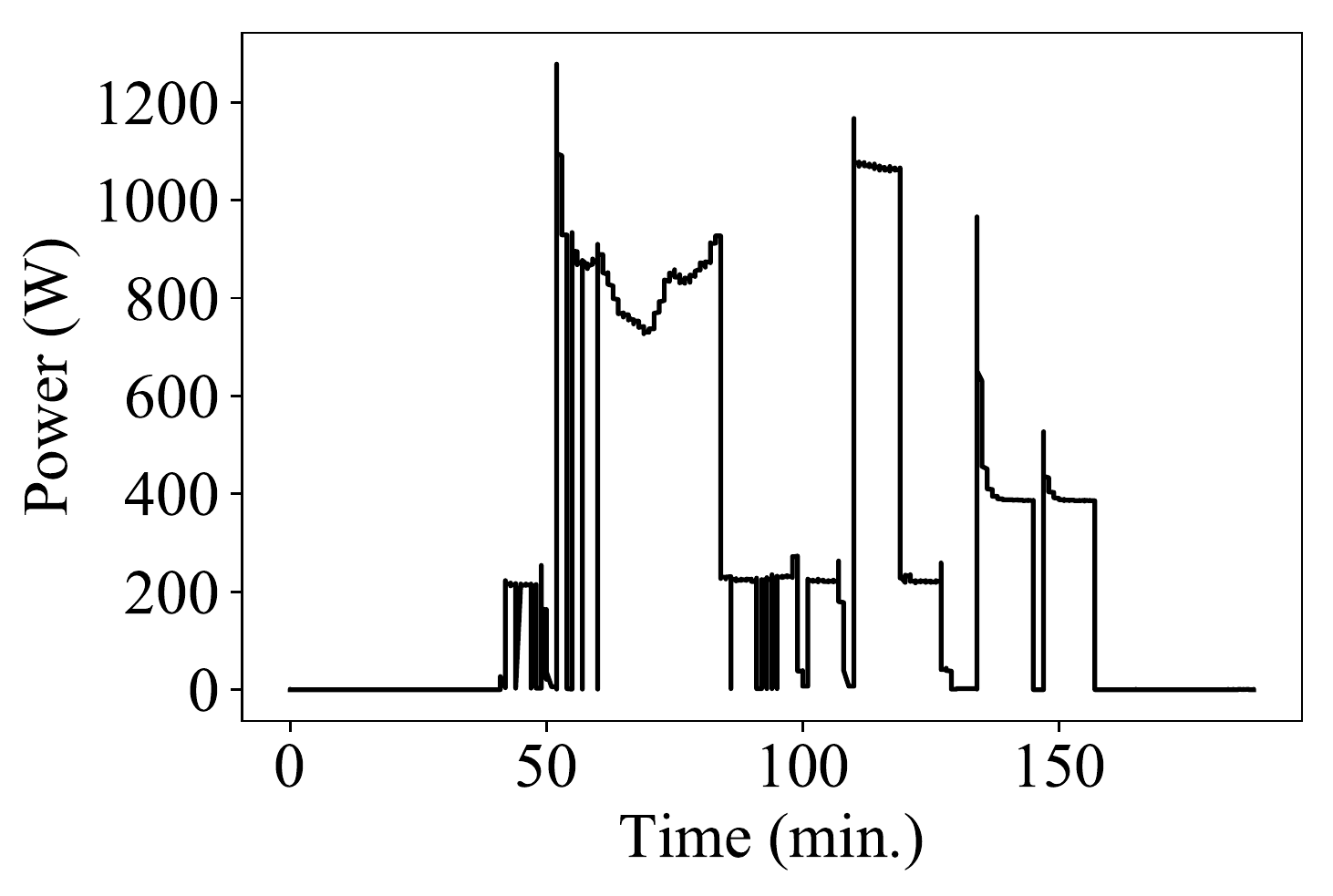}
    \caption*{(a) Dish washer} 
  \end{subfigure}
  \begin{subfigure}[b]{0.5\linewidth}
    \centering
    \includegraphics[width=\linewidth]{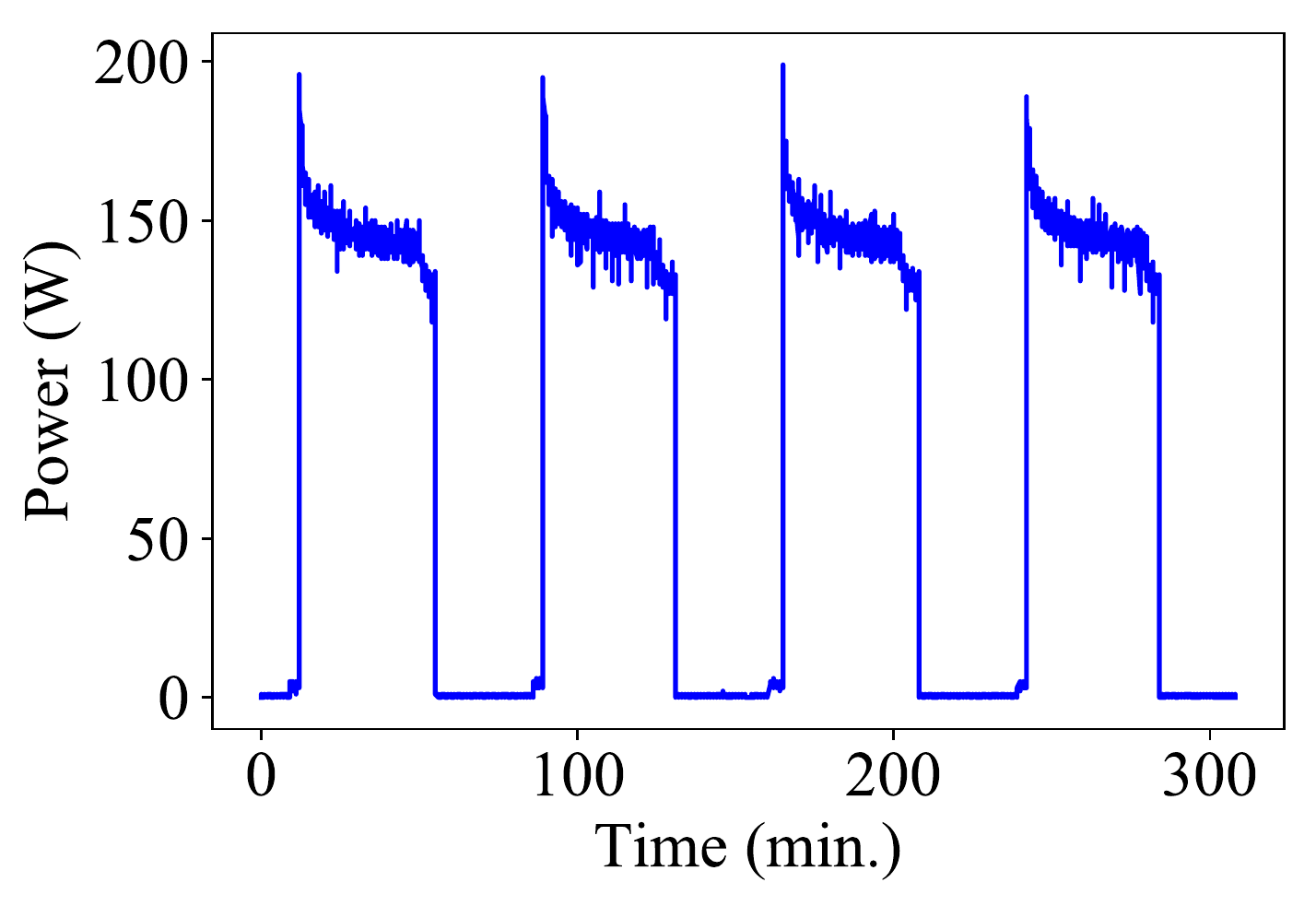}
    \caption*{(b) Fridge} 
  \end{subfigure} 
  \begin{subfigure}[b]{0.5\linewidth}
    \centering
    \includegraphics[width=\linewidth]{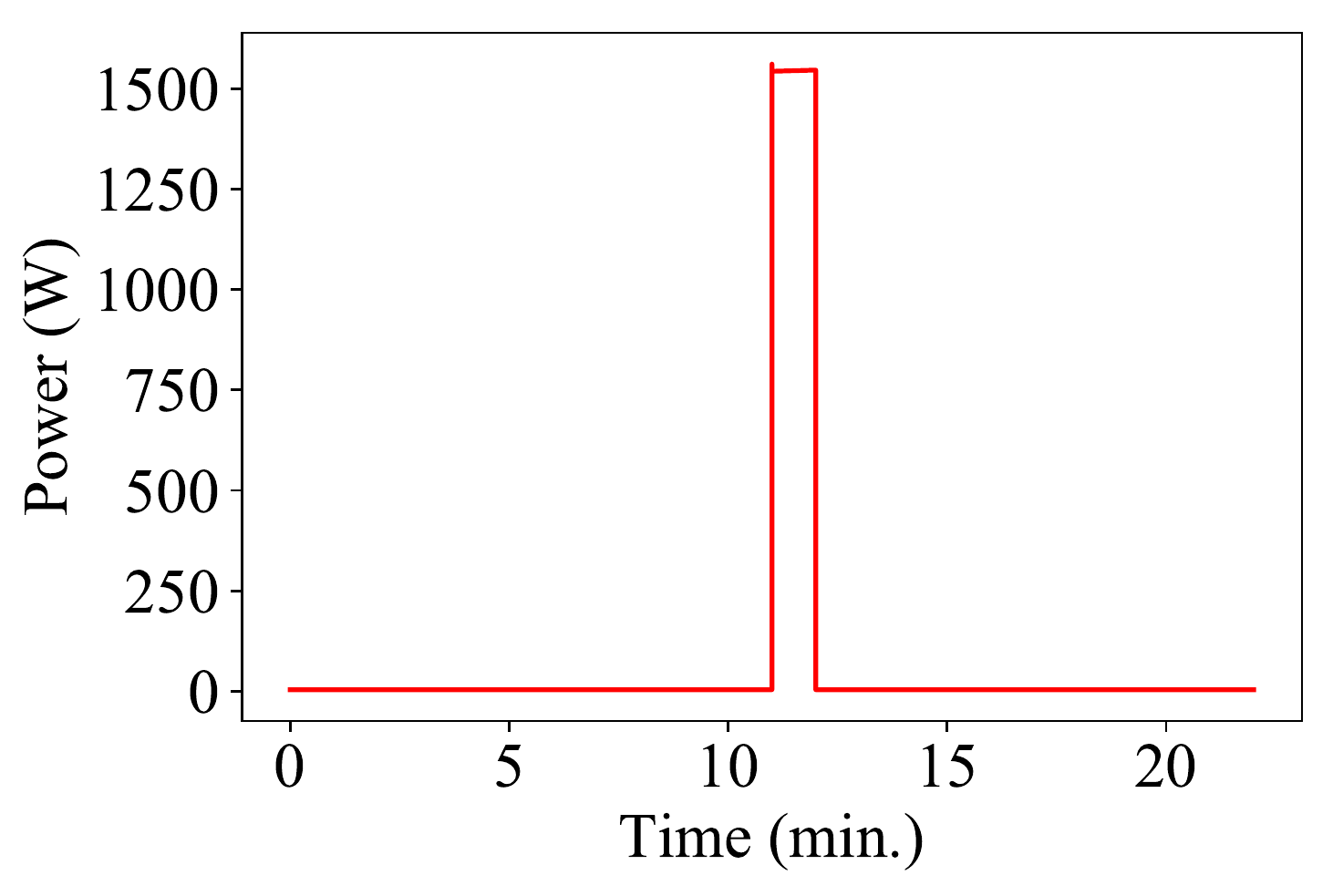}
    \caption*{(c) Microwave} 
  \end{subfigure}
  \begin{subfigure}[b]{0.5\linewidth}
    \centering
    \includegraphics[width=\linewidth]{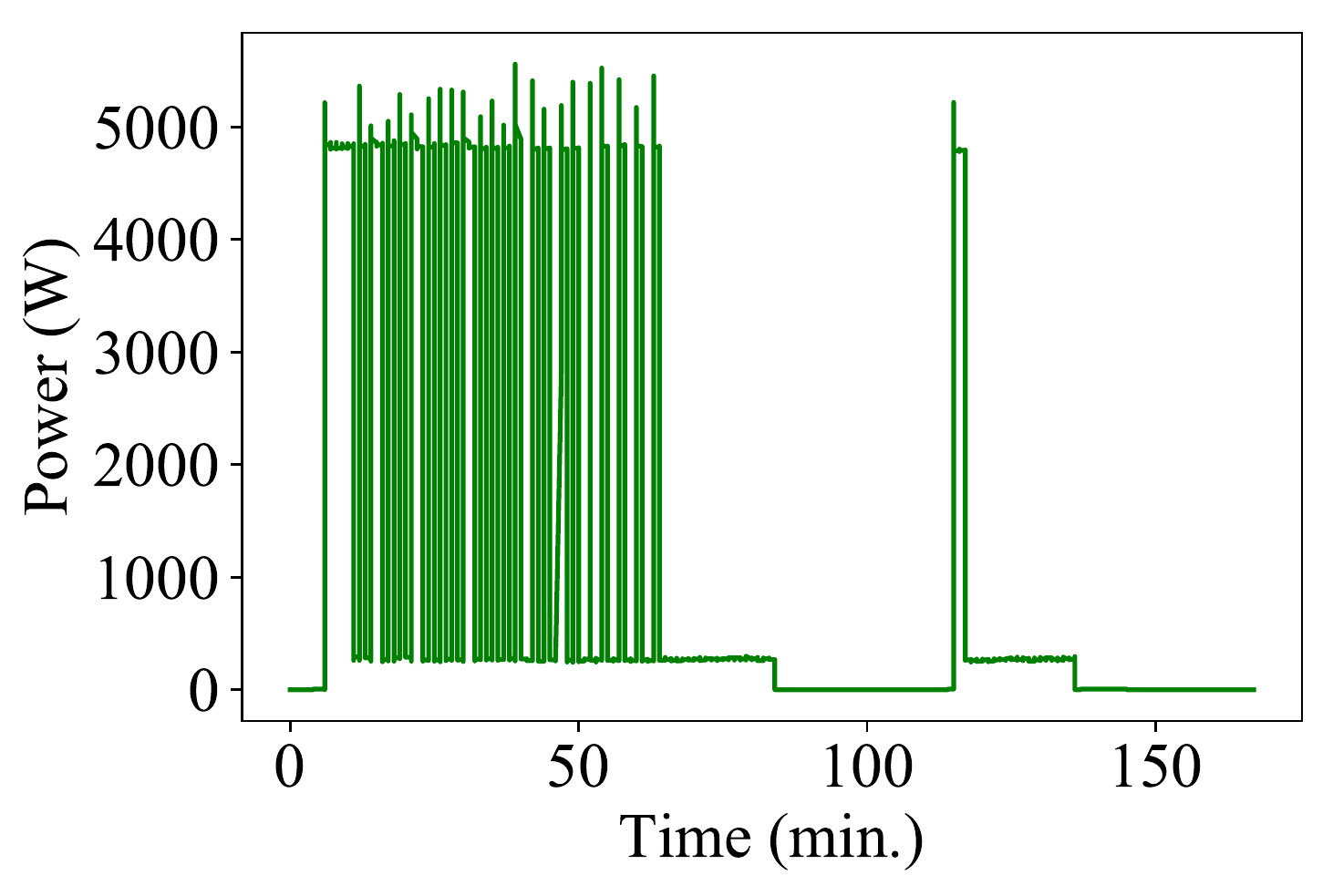}
    \caption*{(d) Washing machine} 
  \end{subfigure} 
  \caption{Snippets of individual appliances in REDD.}\label{snippets}
\end{figure}

In the last few years, deep learning has become a popular approach for NILM \cite{kelly2015neural,huss2015hybrid,zhang2016sequence}. These works show improved NILM performance by applying deep learning algorithms that mainly developed in other fields. However, the previous works do not explicitly exploit the inherent properties of electricity consumption data. In our work, we utilize one of these properties, the on/off state concept of appliances that we can observe in Figure \ref{snippets}. We present subtask gated networks (SGN) that adopt one DNN for regression, and the other DNN for on/off classification. By multiplying regression output with classification probability to form the final estimates, this model outputs the power estimation gated with on/off classification. Through the loss function that explicitly reflects the classification errors, classification subnetwork serves as the on/off based gate. SGN shows 15-30\% improved performance in REDD and UK-DALE on average. Also, we further investigated its variants that reflects standby power and hard gating, which shows additional improvements for some appliances.   

\section{Related Works}
\label{sec:relatedworks}
\subsection{Non-intrusive load monitoring}
The most frequently used approach for NILM is the Factorial Hidden Markov Model (FHMM) \cite{ghahramani1997factorial}
and its variants \cite{kim2011unsupervised,kolter2012approximate,parson2012non,zhong2014signal,shaloudegi2016sdp}.
Kim et al. proposed an energy disaggregation algorithm based on a conditional factorial hidden semi-Markov model,
which exploits additional features related to when and how appliances are used \cite{kim2011unsupervised}.
Kolter and Jaakkola presented an approximate factorial hidden Markov model based on convex optimization for inference,
that aggregates the difference signal and other constraints for aggregated energy data \cite{kolter2012approximate}.
Parson et al. showed that prior models of appliances in FHMM could be applied to real-world energy disaggregation system \cite{parson2012non}.
Zhong et al. incorporated signal aggregate constraints (SACs) into an additive factorial hidden Markov model, which significantly improves the original additive factorial hidden Markov model \cite{zhong2014signal}.
Shaloudegi et al. enhanced Zhong et al.'s algorithms by combining convex semidefinite relaxations randomized rounding and a scalable ADMM optimization algorithm \cite{shaloudegi2016sdp}. 

The latest approaches are based on deep learning.
Jack Kelly et al. tried a variety of deep learning models including Convolutional Neural Networks (CNNs), Recurrent Neural Networks (RNNs), and Denoising Autoencoder (DAE) \cite{kelly2015neural}.
Huss proposed a hybrid energy disaggregation algorithm based on CNNs and a hidden semi-Markov model \cite{huss2015hybrid}.
Zhang et al. suggested a sequence-to-point learning with CNN such that
the single mid-point of a time window is treated as the output of the network instead of the whole sequence of the window \cite{zhang2016sequence}. While showing significant performance improvements, none of the approaches utilize on/off states of appliances with end-to-end training.

\subsection{Multi-task learning}
Our work was inspired by Multi-Task Learning (MTL). In general, multi-task learning learns with multiple tasks that are associated with task-specific loss functions \cite{ruder2017overview}. 
Multi-task learning methods enable models to generalize better for the original task by sharing parameters among the related tasks. Caruana summarizes the goal of MTL as following --- ``MTL improves generalization by leveraging the domain-specific information contained in the training signals of related tasks" \cite{caruana1998multitask}. 

Most multitask learning studies have focused on improving the performance of the main task or all tasks by lower layer parameter sharing among multiple tasks \cite{zhang2014facial,yu2016learning,liu2015representation,cheng2015open}. For example, In the work of Zhang et al., head pose estimation and facial attribute inference tasks are used as the auxiliary tasks, which shares parameters except for the last layer's parameters that participate in each tasks' output. These related tasks improve the facial landmark detection task by learning better parameter values \cite{zhang2014facial}. 

The proposed architecture in this paper, subtask gated networks, can be interpreted as multi-task learning with an auxiliary task of classification. However, there is a major difference in the way subtask output is used. In our work, the auxiliary task output is used as a multiplication unit to calculate the final output of the main task, and this is beyond parameter sharings that indirectly affect the main task through the loss functions. 

\subsection{Gating mechanism} 
Gating mechanism in deep learning means softly selecting one of two or more components using sigmoid or softmax function. For example, sigmoid has been widely used to determine whether a hidden state should be memorized or not in recurrent neural networks \cite{hochreiter1997long,gers2000recurrent,cho2014properties}. As another example, in the mixture of experts \cite{jacobs1991adaptive}, softmax has been used to decide which expert to use for each input region. However, in these examples, gating networks cannot explicitly learn from gating task itself. Instead, these examples use EM algorithm or backpropagation with the loss function related with the final output, not the gating itself. In our subtask gated network, we added a sort of `gating loss' using auxiliary task label that is available, which makes our model learn from on/off gating task directly.


\section{Preliminary}
\label{sec:preliminary}

\subsection{Problem formulation of energy disaggregation}Given the aggregate power consumption for time $T$ periods as $\mathbf{x} = (x_{1}, x_{2}, \cdots,  x_{T})$, let $\mathbf{y}^{i} = (y^{i}_{1}, y^{i}_{2}, \cdots, y^{i}_{T})$ denote the power consumption of $i$-th appliance in the house. At each time step, the aggregate power consumption could be represented as the summation of individual power consumptions as follows
\begin{equation} \label{eq:aggregation-model}
x_{t} = \sum_{i}y^{i}_{t} + \epsilon_{t},
\end{equation}
where $\epsilon_{t}$ is assumed Gaussian noise with zero mean and variance $\sigma_{t}^{2}$. Suppose we are only interested in I appliances, which are widely used household appliances in most of households. 
Then, the power consumption from unknown appliances can be represented $\mathbf{u} = (u_{1}, u_{2}, \cdots, u_{T})$, and (\ref{eq:aggregation-model}) can be reformulated as follows
\begin{equation} \label{eq:aggregation-model-interest}
x_{t} = \sum_{i}^{I}y^{i}_{t} + u_{t} + \epsilon_{t}.
\end{equation}
The disaggregation problem is then to estimate the power consumption sequences of appliances $\mathbf{y}^{1}, \mathbf{y}^{2}, \cdots, \mathbf{y}^{I}$ from $\mathbf{x}$.

\subsection{Deep neural networks for energy disaggregation}
In previous studies \cite{kelly2015neural,zhang2016sequence},
deep neural networks for nonlinear regression are used to estimate the power consumption of individual appliances
from the sequence of aggregated power consumptions.
For practical reasons, deep neural networks use partial sequences
$\mathbf{x}_{t,s} := (x_t, \cdots, x_{t+s-1})$ and $\mathbf{y}^{i}_{t,s} := (y^{i}_t, \cdots, y^{i}_{t+s-1})$
starting at $t$ with length $s$ as input and output respectively,
rather than the whole sequences of $\mathbf{x}$ and $\mathbf{y}^{i}$.
To avoid the loss of context information,
we can consider additional windows of length $w$ on both end sides, only for input.
To be precise,
$\tilde{\mathbf{x}}_{t,s,w} := (x_{t-w}, \cdots, x_{t+s+w-1})$ as the input and $\mathbf{y}^{i}_{t,s}$ as the output.
Since the sequence length $s$ and the additional window length $w$ are fixed throughout this paper,
we will omit the subscripts $s$ and $w$ on notations for simplicity.
In our work, we denote the appliance power estimation model $f^{i}_{power}:\mathbb{R}_{+}^{s+2w} \rightarrow \mathbb{R}_{+}^{s}$ for an individual appliance is defined as

\begin{equation} 
\hat{\mathbf{p}}^{i}_{t} = f^{i}_{power}(\tilde{\mathbf{x}}_{t}).
\label{eq:disaggregation-power}
\end{equation} 

\textbf{Architectures}
Although RNN architectures are widely used for sequence modeling \cite{sutskever2014sequence},
CNN is also an attractive solution for energy disaggregation problems and shows better performance than the RNN architectures \cite{kelly2015neural,huss2015hybrid,zhang2016sequence}.
Recently, an empirical study showed that CNNs can outperform RNNs across a diverse range of sequence modeling tasks and data sets \cite{bai2018empirical}. 
When some of RNNs were tried for our problem, they showed worse performance than CNNs even with a longer training time.
Therefore, RNNs are excluded in this study and we only consider the CNN model used in Zhang's paper \cite{zhang2016sequence}.


\section{Subtask Gated Networks (SGN)}
\label{sec:main-algorithm}

\begin{figure}

  \centering
    \includegraphics[width=0.5\textwidth]{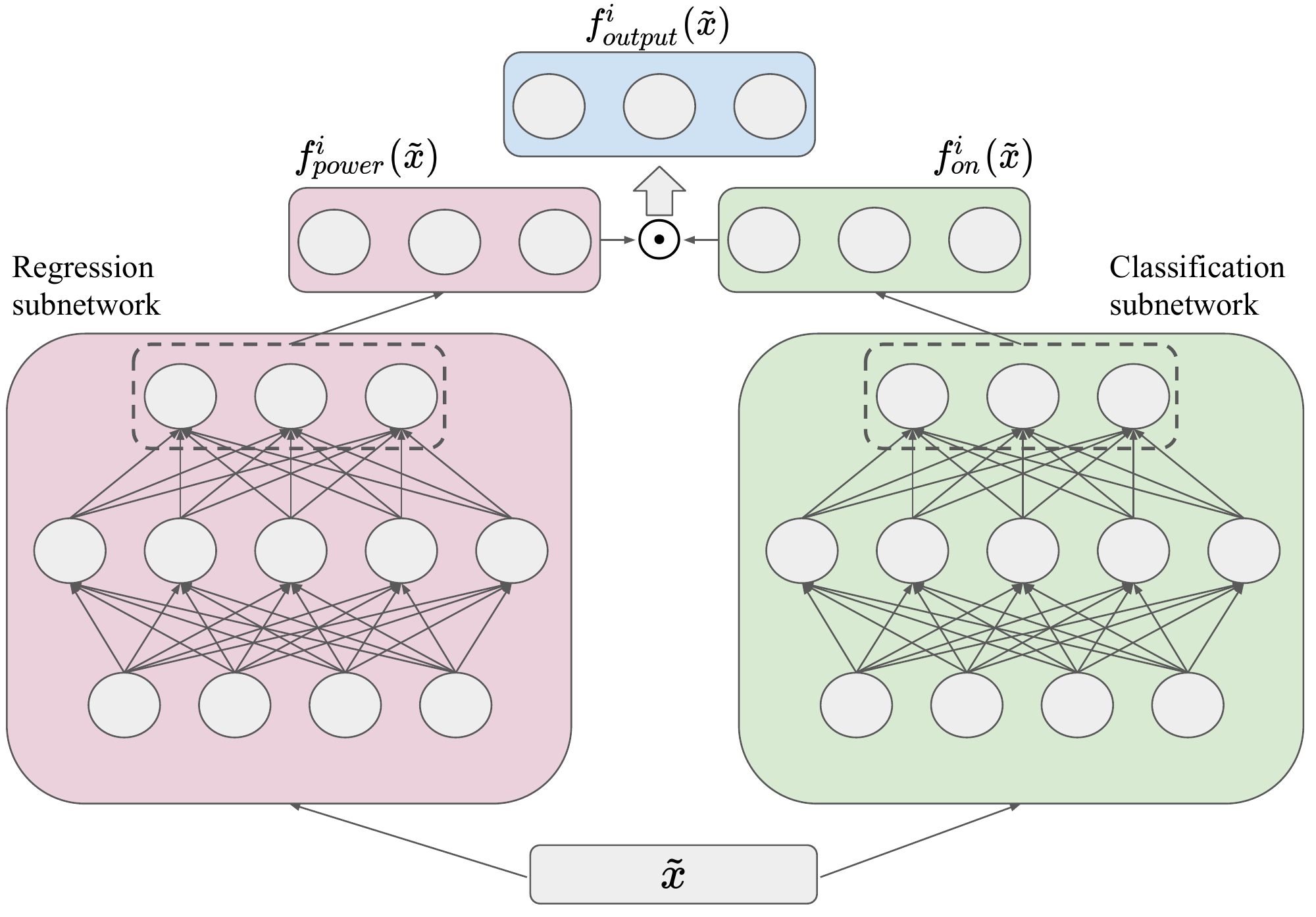}
    \caption{A generic subtask gated network.} \label{generic_SGN}
\end{figure}

\subsection{Auxiliary classification subtask for NILM}
Our general framework uses the auxiliary classification subtask as the gating unit of the main regression task as shown in Figure \ref{generic_SGN}. Using the subtask, we can exploit the on/off state in an explicit way. To be precise, we can formulate on/off classification as a subtask for non-intrusive load monitoring.

Let $\mathbf{o}^{i} = (o^{i}_{1}, o^{i}_{2}, \cdots, o^{i}_{T})$ be the state sequence of appliance $i$ where $o^{i}_{t} \in \{0,1\}$ represents the on/off state of appliance $i$ at time $t$:

\begin{equation} \label{on-off-labelling}
   o^{i}_{t}= 
    \begin{cases}
        1, & \text{if $y^{i}_{t} >$ threshold,} \\
        0, & \text{otherwise.}
    \end{cases}
\end{equation}

Again, we use the notation $\mathbf{o}^{i}_{t,s} := (o^{i}_t, \cdots, o^{i}_{t+s-1})$ for partial sequence.
The subscript $s$ will be omitted as long as the sequence length $s$ is fixed.
And then, we define a neural network
$f^{i}_{on}:\mathbb{R}_{+}^{s+2w} \rightarrow [0,1]^{s}$
for on/off state classification of appliance $i$, which is the mapping:
\begin{equation} \label{eq:classification-onoff-multiple}
\hat{\mathbf{o}}^{i}_{t} = f^{i}_{on}(\tilde{\mathbf{x}_{t}}).
\end{equation}
We want to emphasize that each element in $\hat{\mathbf{o}}^{i}_{t}$
means a probability that an appliance has `on' state.

We define $f^{i}_{output}$, the final output of our structure, as
\begin{equation} \label{eq:joint-disaggregation-classifiction}
f^{i}_{output}(\tilde{\mathbf{x}}_{t}) = f^{i}_{power}(\tilde{\mathbf{x}}_{t}) \odot f^{i}_{on}(\tilde{\mathbf{x}}_{t})
\end{equation}
where $\odot$ represents the element-wise multiplication.
Then, this architecture naturally induces the following loss functions:

\begin{subequations}
\begin{align} \label{eq:loss-output}
\mathcal{L}_{output}^{i} = \frac{1}{T}
\sum_{t=1}^{T} ( y^{i}_{t}-\hat{p}^{i}_{t} \hat{o}^{i}_{t})^{2},
\end{align}
\begin{align} \label{eq:loss-power}
\mathcal{L}_{power}^{i} = \frac{1}{T}
\sum_{t=1}^{T} (y^{i}_{t}-\hat{p}^{i}_{t})^{2},
\end{align}
\begin{align} \label{eq:loss-on}
\begin{split}
\mathcal{L}_{on}^{i} =
- \sum_{t=1}^{T}
\Big(
o^{i}_{t} \log \hat{o}^{i}_{t}
+ \left(1-o^{i}_{t}\right) \log (1 - \hat{o}^{i}_{t})
\Big).
\end{split}
\end{align}
\end{subequations}
Note that $\mathcal{L}^{i}_{output}$ and $\mathcal{L}^{i}_{power}$ are the mean squared error loss in the whole network and the regression subnetwork for the power estimation task, respectively. $\mathcal{L}^{i}_{power}$ is the loss function used in literatures \cite{kelly2015neural,zhang2016sequence}. $\mathcal{L}^{i}_{on}$ is the sigmoid cross entropy loss in the classification subnetwork.

\subsection{Loss function}
In our work, we use the following loss function for the joint optimization.
\begin{equation}
\mathcal{L} = \mathcal{L}^{i}_{output} + \mathcal{L}^{i}_{on}.
\end{equation}
This loss function represents the sum of the overall network loss and the classification subnetwork loss.
Our model can be trained using $\mathcal{L}^{i}_{output}$ only, but, $\mathcal{L}^{i}_{on}$ directly instill on/off state information to the classification subnetwork. It induces the classification subnetwork to operate as an on/off gate and using $\mathcal{L}^{i}_{output}$ and $\mathcal{L}^{i}_{on}$ shows better performance.

The loss function and its gradient for the whole network are:
\begin{align*}
\mathcal{L}_{output}^{i}
&= \frac{1}{T} \sum_{t}^{T} (y^{i}_{t}-\hat{p}_{t}^{i} \hat{o}^{i}_{t})^{2}, \\
\frac{\partial \mathcal{L}_{output}^{i}}{\partial \hat{y}_{j}^{i}}
&= -\frac{2}{T} \hat{o}_{j}^{i} \left( y_{j}^{i} - \hat{p}_{j}^{i}\hat{o}_{j}^{i} \right), \\
\frac{\partial \mathcal{L}_{output}^{i}}{\partial \hat{o}_{j}^{i}}
&= -\frac{2}{T} \hat{y}_{j}^{i} \left( y_{j}^{i} - \hat{p}_{j}^{i} \hat{o}_{j}^{i} \right).
\end{align*}

Above equations show that the regression subnetwork learns only when the classification subnetwork classifies the state of the target appliance as `on' state.
However, if the classification subnetwork is saturated as zero for the input, the power estimation subnetwork cannot learn because gradients are also zeros. For that case, minimizing $\partial \mathcal{L}_{output}^{i}/\partial \hat{o}_{j}^{i}$ can revive the classification network later if the regression subnetwork outputs non-zero values.

\subsection{Variants of subtask gated networks for NILM}
Here, we consider several options to enhance the performance of SGN for NILM problem.
\subsubsection{SGN with standby power (SGN-sp)}
\begin{table}[h]
  \centering
  \begin{tabular}{c | c | c }
    \toprule
     & REDD & UK-DALE \\ 
    \midrule
    Dish washer & 0.27 & 0.61 \\
    Fridge & 3.82 & 3.02 \\
    Kettle & - & 1.10 \\
    Microwave & 3.66 & 0.50 \\
    Washing machine & 1.14 & 2.76 \\
    \bottomrule
  \end{tabular}
  \caption{Mean standby power (W) of appliances.}
  \label{standby-power}
\end{table}

Appliances may have non-zero power consumption even when it is in `off' state. By analyzing data, we have easily identified standby power as shown in Table \ref{standby-power}. They may seem small, but they are not negligible because the `off' state takes up a significant portion of the time. 
Because SGN has the assumption that an appliance power consumption is zero when it is in `off' state, a slight modification is required to take into account standby power.
Considering that standby power is a fixed value when the appliance is in `off' state, the formulation can be simply modified as follows.
\begin{equation} 
f^{i}_{output}(\tilde{\mathbf{x}}_{t}) = f^{i}_{power}(\tilde{\mathbf{x}}_{t}) \odot f^{i}_{on}(\tilde{\mathbf{x}}_{t}) + (\mathbbm{1}- f^{i}_{on}(\tilde{\mathbf{x}}_{t}))b,
\end{equation}
where $b$ is a learnable scalar.

\subsubsection{Hard SGN}
Instead of multiplying the power estimation value by the probability output of classification subnetwork as it is, multiplying the power estimation by 0 or 1 depending on the output of classification subnetwork may be more compliant to the meaning of the on/off gating. The hard gating can be made simply by applying a condition function $g(x)$ such that \\
$$g(x)= 
\begin{cases}
    1, & \text{if } x \ge 0.5, \\
    0, & \text{otherwise.}
\end{cases}
$$
The final output is modified as follows.
\begin{equation} 
f^{i}_{output}(\tilde{\mathbf{x}}_{t}) = f^{i}_{power}(\tilde{\mathbf{x}}_{t}) \odot g(f^{i}_{on}(\tilde{\mathbf{x}}_{t})).
\end{equation}

We can also derive \textbf{Hard SGN-sp} model straightforwardly with the modified output formulation as follows.

\begin{equation} 
f^{i}_{output}(\tilde{\mathbf{x}}_{t}) = f^{i}_{power}(\tilde{\mathbf{x}}_{t}) \odot g(f^{i}_{on}(\tilde{\mathbf{x}}_{t})) + (\mathbbm{1}- g(f^{i}_{on}(\tilde{\mathbf{x}}_{t})))b.
\end{equation}

Note that $\mathcal{L}_{on}$ is still based on $\hat{o}^{i}_{t}$ in Hard SGN and Hard SGN-sp, not $g(\hat{o}^{i}_{t})$.


\section{Experiments}\label{Experiments}

\subsection{Experiment settings}
\textbf{Datasets}
We evaluate our proposed methods on the two real-world datasets, REDD \cite{kolter2011redd} and UK-DALE \cite{kelly2014uk}.
The REDD dataset has the data for six US houses, and the UK-DALE dataset has the electricity usage data for five UK houses respectively.

In REDD, the aggregate power consumptions were recorded in every 1 second, and the appliance power consumptions were recorded in every 3 seconds over various durations.
The dataset contains the measurements of the aggregate power consumption and 10-25 types of appliances. 
However, we only consider microwave, fridge, dish washer and washing machine as in the previous work \cite{zhang2016sequence}.
We used the data of house 2--6 as the training set, and house 1 as the test set.

In UK-DALE, all data were recorded in every 6 seconds from November 2012 to January 2015. The dataset contains the main aggregated power consumption and measurements of 4-54 appliances. We only consider kettle, microwave, fridge, dish washer and washing machine in our experiments.  
For evaluation, we used house 1 and 3--5 for training, and house 2 for testing as in the previous work \cite{zhang2016sequence}.
We only used the last week data that was published after preprocessing\footnote{http://jack-kelly.com/files/neuralnilm/NeuralNILM\_data.zip}.

\begin{table*}[b]
  \footnotesize
  \centering
  \begin{tabular}{cc*{4}{c}{c}{c}}
    \toprule
    \multirow{2}{*}{Metric} & \multirow{2}{*}{Model} & Dish & \multirow{2}{*}{Fridge} & \multirow{2}{*}{Microwave} & Washing & \multirow{2}{*}{Average} & 
    Average \\
     & & washer & & & machine & & improvement\\ 
     
    \midrule
    \multirow{8}{*}{MAE}    & FHMM \cite{batra2014nilmtk}  & 101.30 & 98.67 & 87.00 & 66.76 & 88.43 & - \\  
                            & DAE \cite{kelly2015neural} & 29.38 & 76.62 & 21.31 & 31.35 & 39.66 & - \\ 
                            & Seq2Seq \cite{zhang2016sequence}& 27.07 & 26.03 & 16.57 & 22.72 & 23.10 & 0.00 \%\\
                            \cmidrule{2-8}
                            & SGN  & \cellcolor[gray]{0.9}\textbf{14.97} & \cellcolor[gray]{0.9}23.89 & 17.52 & \cellcolor[gray]{0.9}20.07 & \cellcolor[gray]{0.9}19.09 & 17.34 \%\\
                            & SGN-sp & \cellcolor[gray]{0.9}15.96 & \cellcolor[gray]{0.9}22.89 & \cellcolor[gray]{0.9}\textbf{15.98} & \cellcolor[gray]{0.9}20.61 & \cellcolor[gray]{0.9}\textbf{18.86} & 18.36 \% \\
                            & Hard SGN & \cellcolor[gray]{0.9}21.27 & \cellcolor[gray]{0.9}24.45  & 17.38  & \cellcolor[gray]{0.9}\textbf{18.24}  & \cellcolor[gray]{0.9}20.33 & 11.97 \% \\
                            & Hard SGN-sp & \cellcolor[gray]{0.9}24.29 & \cellcolor[gray]{0.9}\textbf{22.86}  & 17.16  & \cellcolor[gray]{0.9}21.94  & \cellcolor[gray]{0.9}21.56 & 6.67 \% \\
    \midrule           
    
    \multirow{8}{*}{$SAE_{\delta}$}    & FHMM  & 93.64 & 46.73 & 65.03 & 58.77 & 66.04 & - \\  
                            & DAE & 29.21 & 20.48 & 17.86 & 27.64 & 23.80 & -\\ 
                            & Seq2Seq & 26.93 & 11.67 & 11.43 & 16.82 & 16.71 & 0.00 \%\\
                            \cmidrule{2-8}
                            & SGN  & \cellcolor[gray]{0.9}\textbf{11.74} & \cellcolor[gray]{0.9}\textbf{10.62}  & 14.84  & \cellcolor[gray]{0.9}10.70  & \cellcolor[gray]{0.9}\textbf{11.97} & 28.34 \% \\
                            & SGN-sp & \cellcolor[gray]{0.9}12.07 & 12.26 & \cellcolor[gray]{0.9}\textbf{11.31} & \cellcolor[gray]{0.9}13.28 & \cellcolor[gray]{0.9}12.23 & 26.81 \% \\
                            & Hard SGN  & \cellcolor[gray]{0.9}20.72 & 14.51  & 16.53  & \cellcolor[gray]{0.9}\textbf{8.47}  & \cellcolor[gray]{0.9}12.40 & 25.77 \% \\
                            & Hard SGN-sp & 27.63 & \cellcolor[gray]{0.9}10.99  & 15.52  & \cellcolor[gray]{0.9}15.66  & \cellcolor[gray]{0.9}13.31 & 20.37 \% \\
                        
    \bottomrule
  \end{tabular}
  \caption{Experiment results for REDD data.}
  \label{redd-table}
\end{table*}

\textbf{Data Preprocessing}
Since the UK-DALE data set has been preprocessed by Jack Kelly \cite{kelly2015neural}, we used it as it is. 
For REDD dataset, we preprocessed with the following procedure to handle missing values.
First, we split the sequence so that the duration of missing values in subsequence is less than 20 seconds.
Second, we filled the missing values in each subsequence by the backward filling method.
Finally, we only used the subsequences with more than one-day duration. 
For both UK-DALE and REDD data, the aggregate power consumptions and appliances' power consumptions were divided by standard deviations of the aggregate power consumptions.

\begin{figure}[h!]
    \centering
    \includegraphics[width=0.41\textwidth]{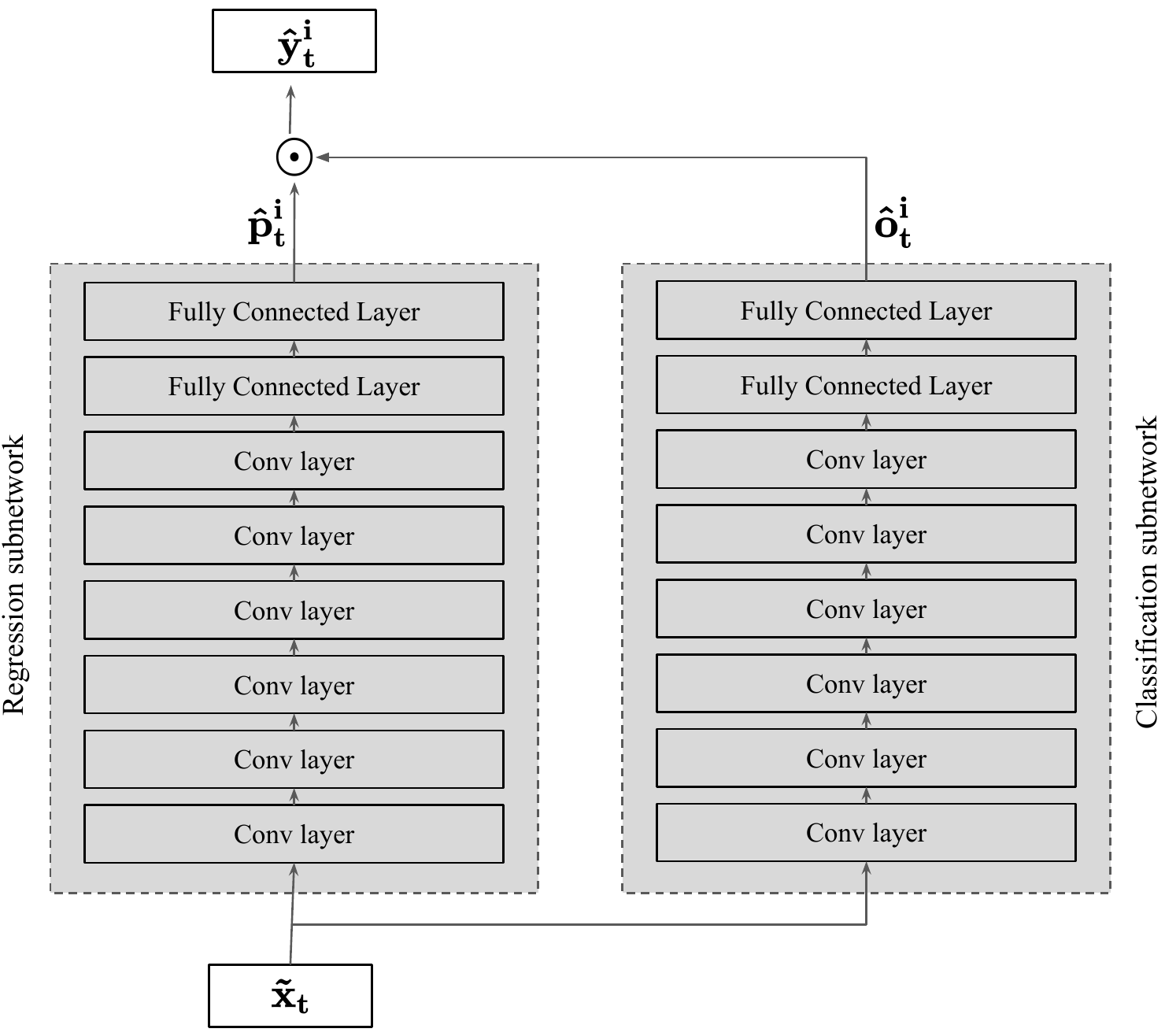}
\caption{SGN architecture used in our experiments. }
\label{SGN_in_experiments}
\end{figure}

\textbf{Neural Net Training Details}
As a baseline, the performance of FHMM was evaluated using FHMM implementations in NILMTK, an open source toolkit for analysis on energy disaggregation \cite{batra2014nilmtk}. 
Jack Kelly's denoising autoencoder (DAE) and Zhang's CNN (Seq2Seq) were implemented according to the architectures described in each paper \cite{kelly2015neural,zhang2016sequence} for evaluations. Our model architecture is shown in Figure \ref{SGN_in_experiments}. This architecture uses Zhang's Seq2Seq as the two subnetworks because neural network architecture itself is not the main focus of our work. More details of each architecture can be found in the supplementary document.
Our model has the following hyperparameters.
The learning rate is $1.0 \times 10^{-4}$, and the batch size is 16. 
The label for the on/off classification task is generated using 15 watts as the threshold according to equation (\ref{on-off-labelling}).
The DNN models are trained on NVIDIA GTX 1080Ti and implemented using TensorFlow 1.8 package. He initialization \cite{he2015delving} is used for the weights of all neural network architectures.
Data was sliced with additional window size $w$=400 and output sequence length $s$=64 for REDD, $w$=200 and $s$=32 for UK-DALE, in which the input is a sequence of 43.2 minutes and the output is a sequence of 3.2 minutes. We used Adam optimizer \cite{kingma2014adam} for training. 

\textbf{Evaluation Metrics}
In this experiment, MAE (Mean Absolute Error) and $SAE_{\delta}$ (Signal Aggregate Error per time $\delta$ period) were used to evaluate the performance. MAE is a general metric used for regression problems, and it shows how well models perform power consumption estimation in NILM. The $SAE_{\delta}$ represents the average total error over the time period of $\delta$, which compare the sum of ground truth and estimation over the time period of $\delta$. We use the time normalized version of signal aggregate error ($SAE$). Unlike the original SAE in previous works \cite{zhang2016sequence} that considers the power consumption of overall period by summing up, we compare the usage per hour, since this may be a more important concept regarding energy consumption behavior control\footnote{We also use various values of $\delta$ to analyze time period sensitivity on evaluation (Figure \ref{REDD_SAE}).}. Note that $SAE_{\delta}$ is normalized using the time period (an hour), not the ground truth power consumption as the previous works \cite{zhang2016sequence}. Although it makes an apple-to-apple comparison difficult among different appliances, it can prevent outliers caused by close-to-zero denominator and division by zero when the appliance is in the `off' state.

Let $y_t^{i}$ and $\hat{y}_t^{i}$ be the ground truth and the estimated power for appliance $i$ at time $t$, respectively.
The MAE for appliance $i$ is defined as:
$MAE^{i} = \frac{1}{T} \sum_{t=1}^T |y_t^{i} - \hat{y}_t^{i}|.$
$SAE_{\delta}$ for appliance $i$ is defined as:
$SAE_{\delta}^{i} = \frac{1}{T_{\delta}} \sum_{\tau=1}^{T_{\delta}} \frac{1}{N_{\delta}}|r_{\tau}^{i} - \hat{r}_{\tau}^{i}|,$
where $N_{\delta}$ is the number of data points in the time period $\delta$, $T_{\delta}$ is the total time period in time $\delta$ scale, and $T = T_{\delta} \times N_{\delta}$. $r_{\tau}$ and $
\hat{r}_{\tau}$ represents the sum of power consumption in a given period, $r_{\tau}^{i} = \sum_{t=1}^{N_{\delta}}y_{N_{\delta} \tau +t}^{i}$, and $\hat{r}_{\tau}^{i} = \sum_{t=1}^{N_{\delta}}\hat{y}_{N_{\delta} \tau +t}^{i}$.  In our experiments, $N_{\delta}$ is 1200 for REDD, and 600 for UK-DALE, which correspond to the number of data points in an hour.

\subsection{Experiment results }

\begin{table*}[!t]
  \footnotesize
  \centering
  \begin{tabular}{cc*{5}{c}{c}{c}}
    \toprule
    \multirow{2}{*}{Metric} & \multirow{2}{*}{Model} & Dish & \multirow{2}{*}{Fridge} & \multirow{2}{*}{Kettle} & \multirow{2}{*}{Microwave} & Washing & \multirow{2}{*}{Average} & Average \\
     & & washer & & & & machine & & improvement\\ 
    
    \midrule
    \multirow{8}{*}{MAE}    & FHMM  & 48.25 & 60.93 & 38.02 & 43.63 & 67.91 & 51.75 & -\\  
                            & DAE & 27.42 & 19.93 & 12.53 & 16.77 & 18.30 & 18.99 & -\\ 
                            & Seq2Seq & 23.16 & 16.71 & 11.07 & 9.87 & 11.94 & 14.55 & 0.00 \%\\
                            \cmidrule{2-9}
                            & SGN & \cellcolor[gray]{0.9}15.50 & \cellcolor[gray]{0.9}15.79 & \cellcolor[gray]{0.9}7.08 & \cellcolor[gray]{0.9}6.26 & 12.31 & \cellcolor[gray]{0.9}11.39 & 21.72 \% \\
                            & SGN-sp & \cellcolor[gray]{0.9}\textbf{13.49} & \cellcolor[gray]{0.9}15.32 & \cellcolor[gray]{0.9}\textbf{5.96} & \cellcolor[gray]{0.9}8.64 & \cellcolor[gray]{0.9}\textbf{11.04} & \cellcolor[gray]{0.9}\textbf{10.89} & 25.15 \% \\
                            & Hard SGN  & \cellcolor[gray]{0.9}15.74 & \cellcolor[gray]{0.9}16.08  & \cellcolor[gray]{0.9}6.27  & \cellcolor[gray]{0.9}\textbf{4.53} & 12.40 & \cellcolor[gray]{0.9}11.00 & 24.38 \% \\            
                            & Hard SGN-sp & \cellcolor[gray]{0.9}15.15 & \cellcolor[gray]{0.9}\textbf{10.93}  & \cellcolor[gray]{0.9}7.40  & 19.27  & 13.31  & \cellcolor[gray]{0.9}13.21 & 9.21 \% \\
    \midrule
    \multirow{8}{*}{$SAE_{\delta}$}    & FHMM  & 46.04 & 51.90 & 35.41 & 41.52 & 64.15 & 47.80 & - \\  
                            & DAE & 23.10 & 5.99 & 7.78 & 14.58 & 14.36 & 13.16 & -\\ 
                            & Seq2Seq & 17.31 & \textbf{3.74} & 6.85 & 8.98 & \textbf{9.90} & 9.36 & 0.00 \% \\
                            \cmidrule{2-9}
                            & SGN & \cellcolor[gray]{0.9}10.03 & 7.10  & \cellcolor[gray]{0.9}4.81  & \cellcolor[gray]{0.9}5.99  & 10.40 & \cellcolor[gray]{0.9}7.66 & 18.09 \% \\
                            & SGN-sp  & \cellcolor[gray]{0.9}\textbf{8.80} & 7.56 & \cellcolor[gray]{0.9}\textbf{3.99} & \cellcolor[gray]{0.9}7.90 & 10.28 & \cellcolor[gray]{0.9}7.71 & 17.63 \% \\
                            & Hard SGN  & \cellcolor[gray]{0.9}10.54 & 7.21  & \cellcolor[gray]{0.9}4.35  & \cellcolor[gray]{0.9}\textbf{4.45}  & 11.18  & \cellcolor[gray]{0.9}\textbf{7.55} & 19.34 \% \\
                            & Hard SGN-sp & \cellcolor[gray]{0.9}9.04 & 3.75  & \cellcolor[gray]{0.9}5.21  & 17.52  & 13.11  & 9.73 & -3.95 \% \\    
    \bottomrule
  \end{tabular}
  \caption{Experiment results for UK-DALE data.}
  \label{ukdale-table}
\end{table*}

\begin{figure*}[!b] 
  \begin{subfigure}[b]{0.33\linewidth}
    \centering
    \includegraphics[width=\linewidth]{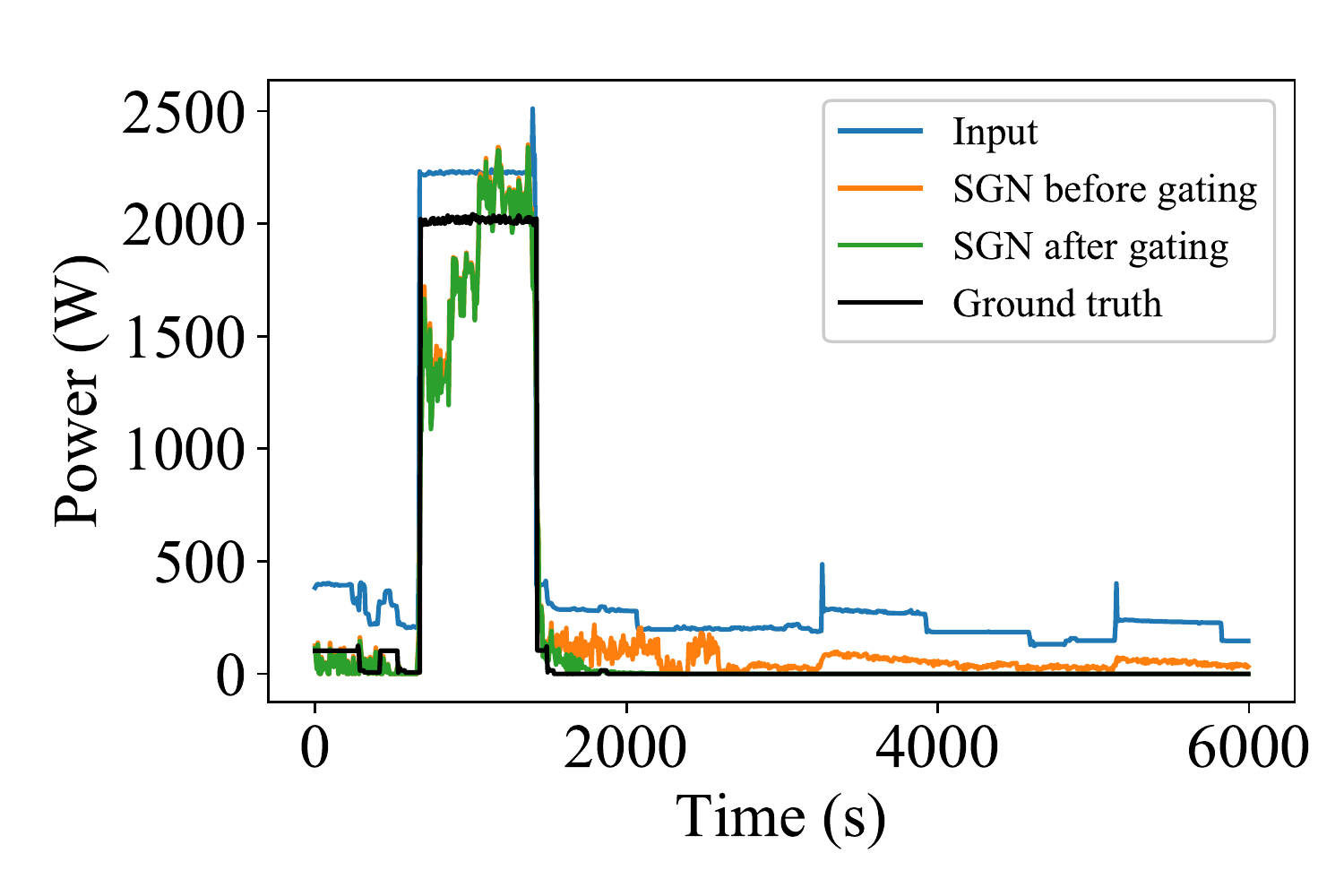}
    \caption*{(a) Dish washer} 
  \end{subfigure}
  \begin{subfigure}[b]{0.33\linewidth}
    \centering
    \includegraphics[width=\linewidth]{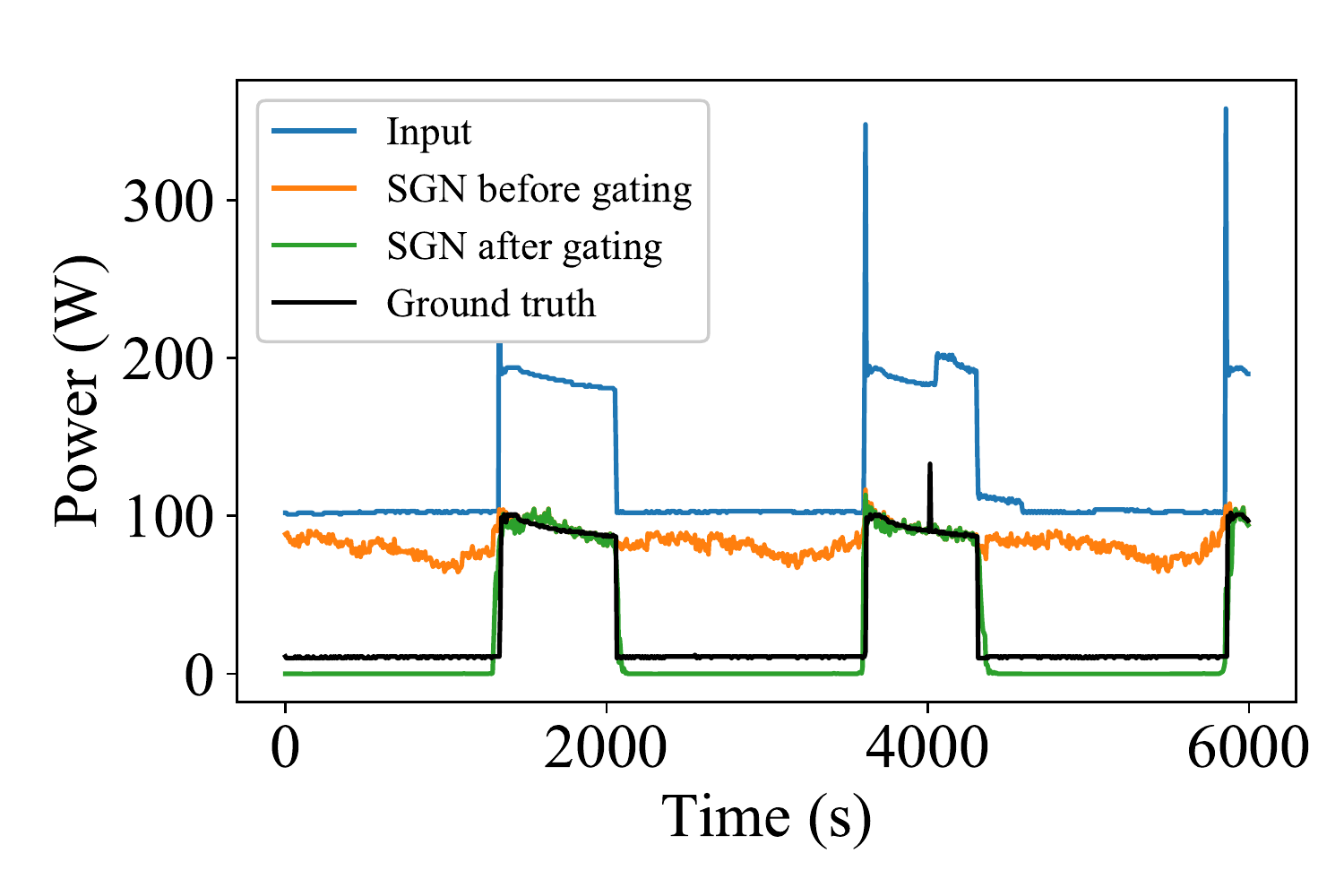}
    \caption*{(b) Fridge} 
  \end{subfigure}
    \begin{subfigure}[b]{0.33\linewidth}
    \centering
    \includegraphics[width=\linewidth]{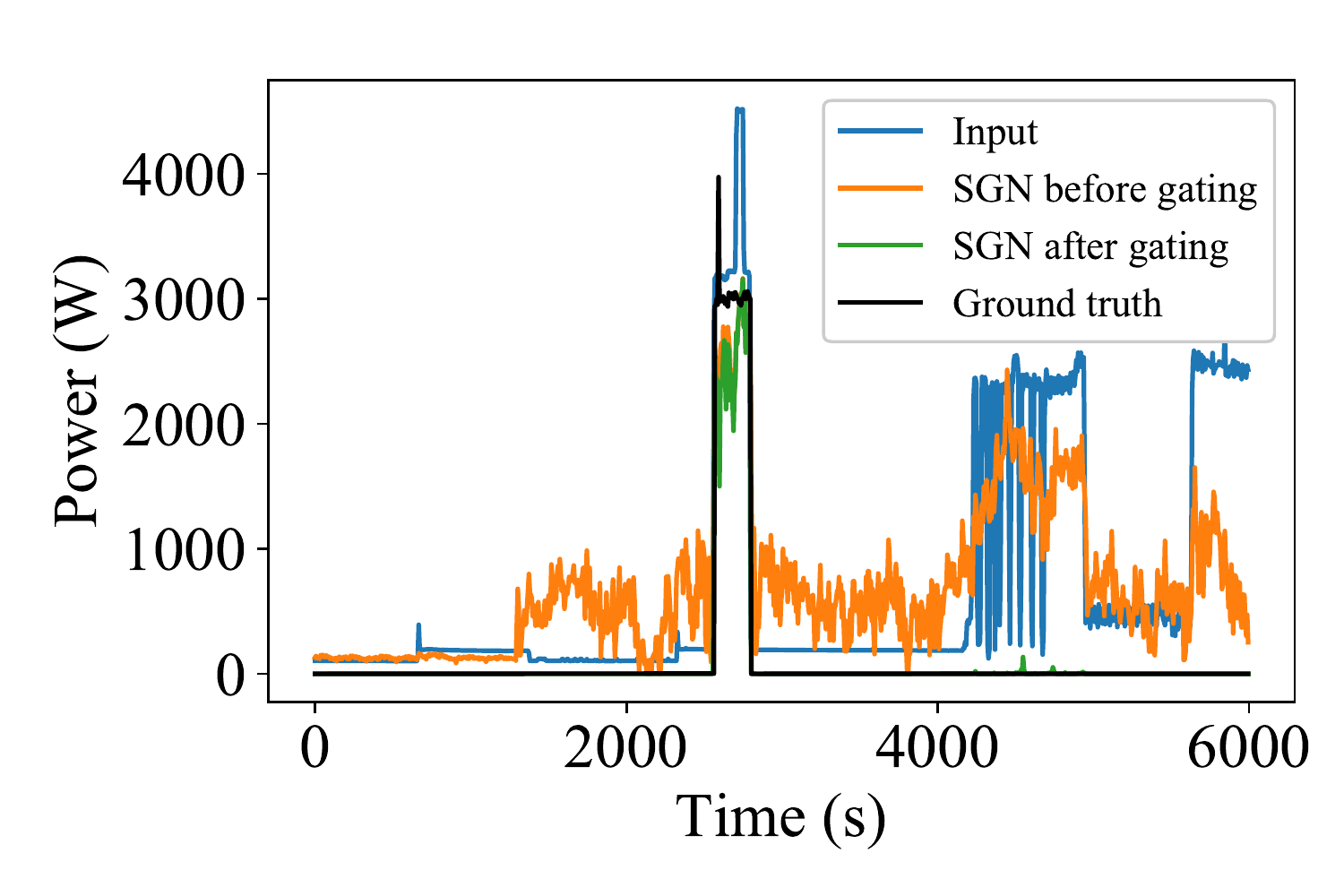}
    \caption*{(c) Kettle} 
  \end{subfigure}
  \caption{Gating examples (UK-DALE).}
  \label{gating-examples}
\end{figure*}

Table \ref{redd-table} and \ref{ukdale-table} show the performance of the previous works and SGN for the REDD and UK-DALE dataset. In the two tables, the bold font denotes the best performing algorithm for each appliance, and the shade denotes that SGN has outperformed the best of the previous works.
For MAE, we can see that the performance is improved for 9 out of 9 cases using SGN-sp, and 7 out of 9 cases using SGN. For SAE, the performance is improved for 6 out of 9 cases using either SGN or SGN-sp. On the average, SGN and SGN-sp show 15-30\% lower error performance compared to the state-of-the-art.
These performance improvements were achieved without fine tuning any of the network architecture, hyperparameters, or the weight between $\mathcal{L}_{output}$ and $\mathcal{L}_{on}$.
Hard SGNs tend to perform worse than SGNs on the average but still perform better than the previous works, and significant improvements were observed for fridges and microwaves. In the real world, we do not need to apply a single model to all the appliances. We can choose from SGN and its variants depending on the appliances type.

\begin{figure*}[h] 
  \begin{subfigure}[b]{0.33\linewidth}
    \centering
    \includegraphics[width=\linewidth]{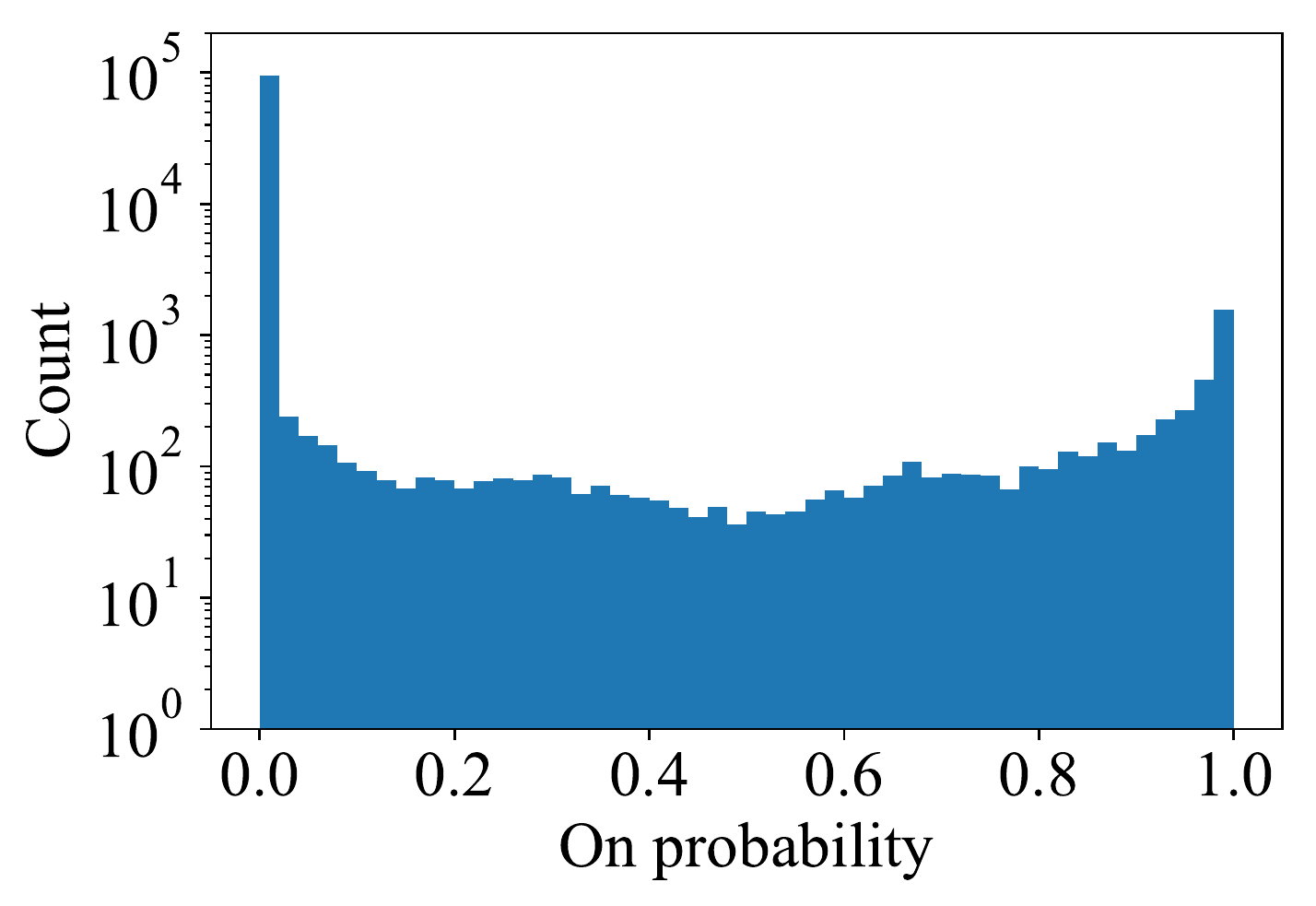}
    \caption*{(a) Dish washer} 
  \end{subfigure}
  \begin{subfigure}[b]{0.33\linewidth}
    \centering
    \includegraphics[width=\linewidth]{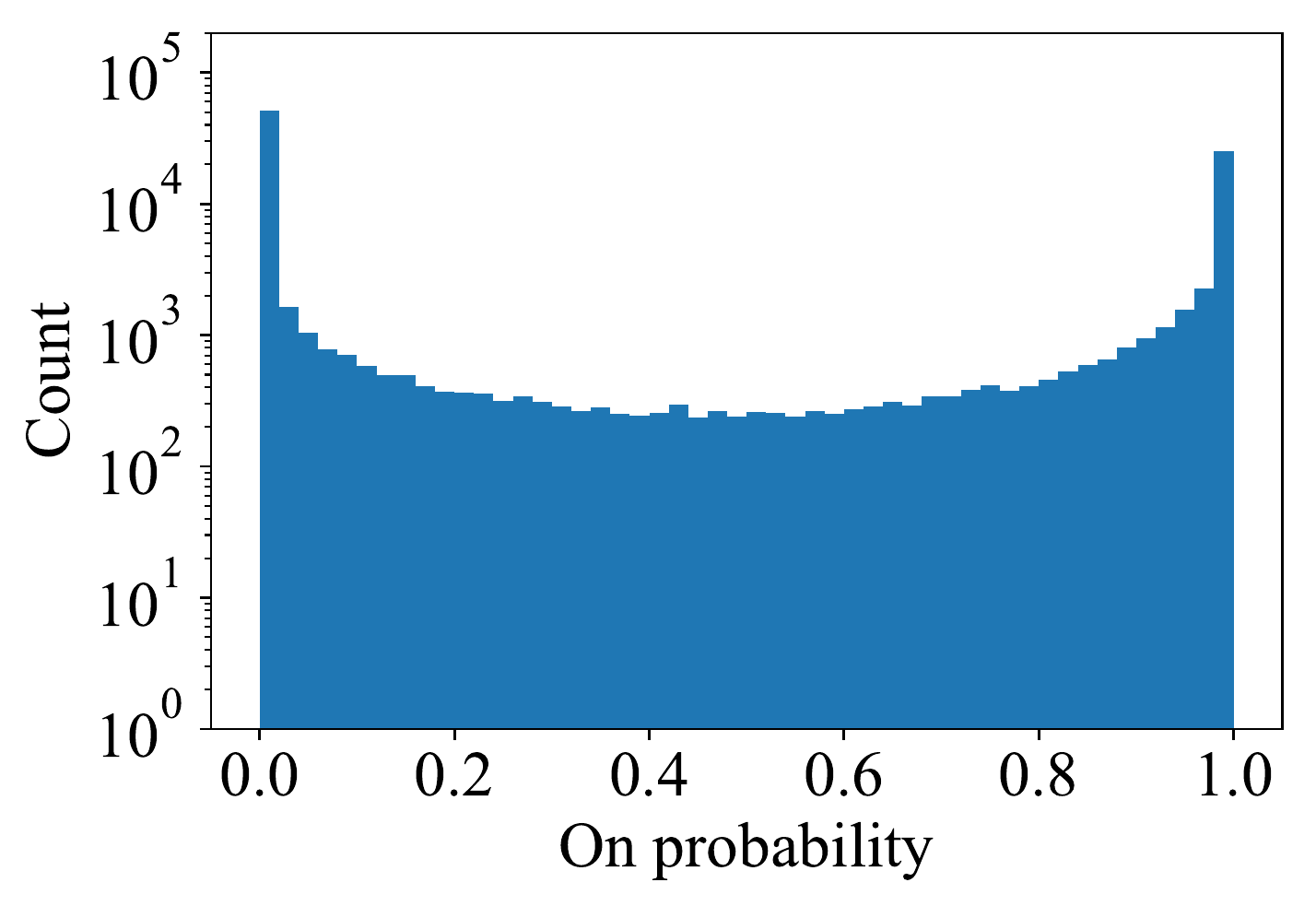}
    \caption*{(b) Fridge} 
  \end{subfigure}
    \begin{subfigure}[b]{0.33\linewidth}
    \centering
    \includegraphics[width=\linewidth]{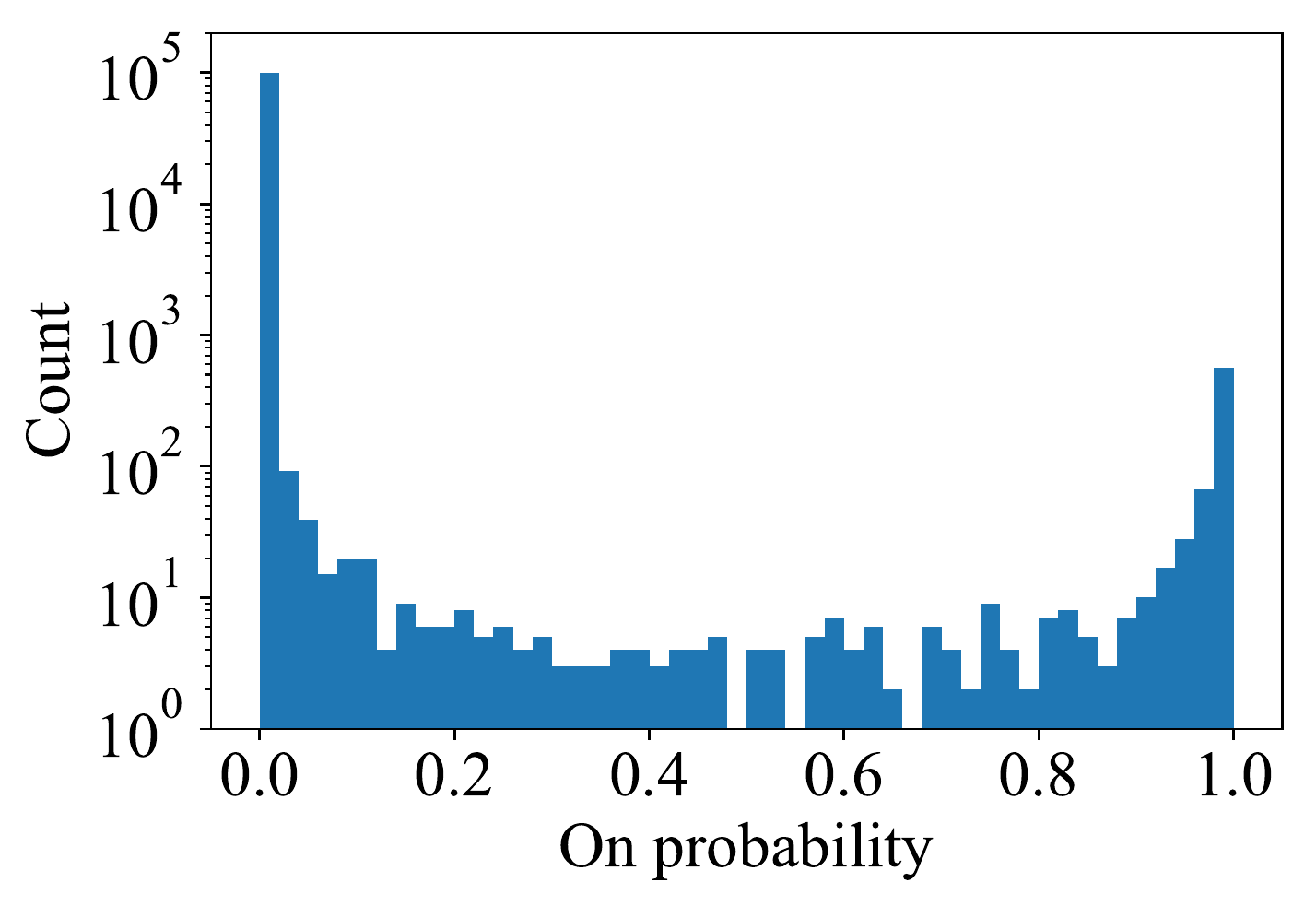}
    \caption*{(c) Kettle} 
  \end{subfigure}
  \caption{Histograms of `on' probability $\hat{o}_{t}^{i}$ (UK-DALE).}
  \label{on-prob}
\end{figure*}

\begin{figure*}[h!] 
\centering
  \begin{subfigure}[t]{0.405\linewidth}
    \centering
    \includegraphics[width=\linewidth]{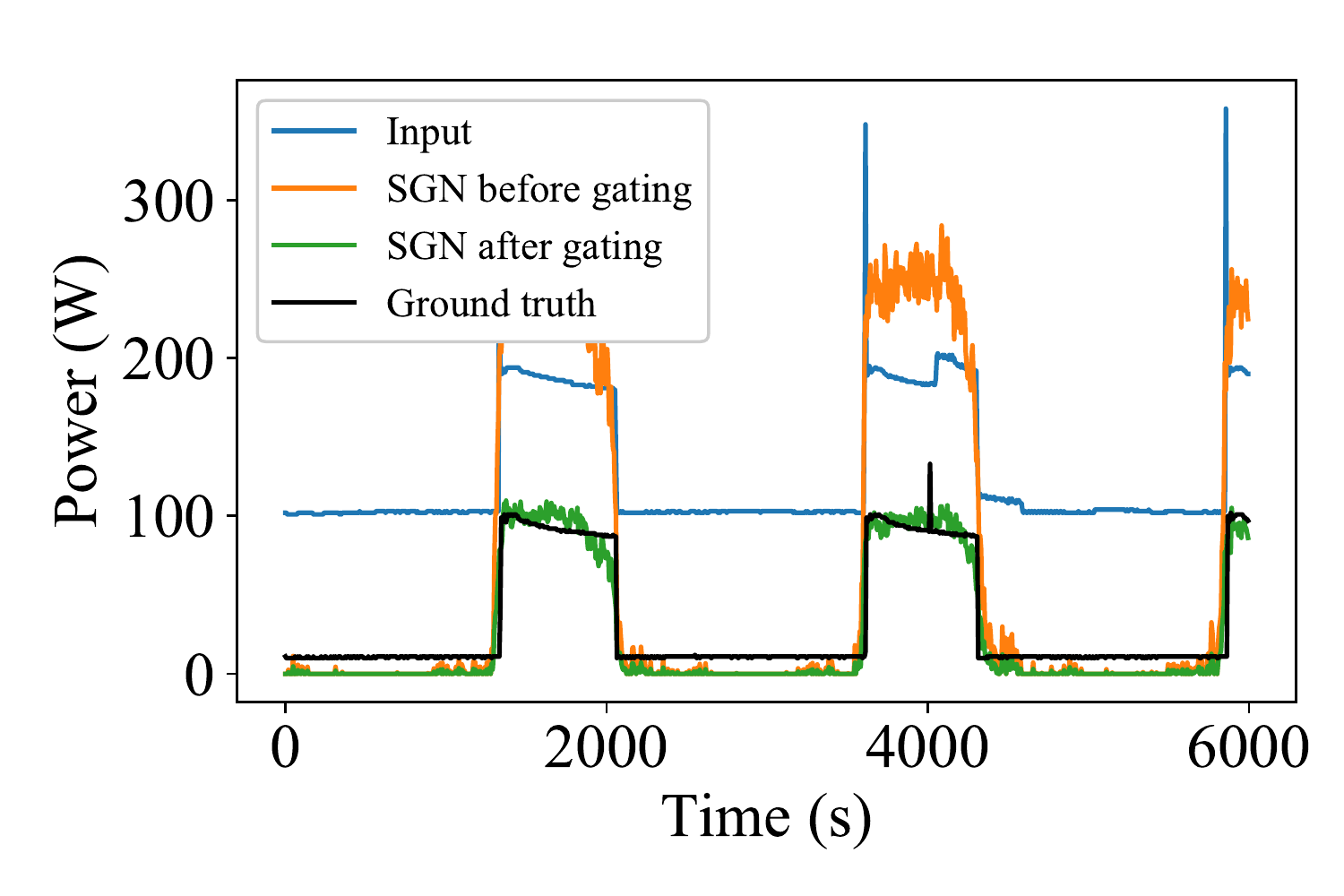}
    \caption*{(a) A gating example of SGN with $\mathcal{L}_{output}$ only} 
  \end{subfigure}
  \hspace{5ex}
  \begin{subfigure}[t]{0.37\linewidth}
    \centering
    \includegraphics[width=\linewidth]{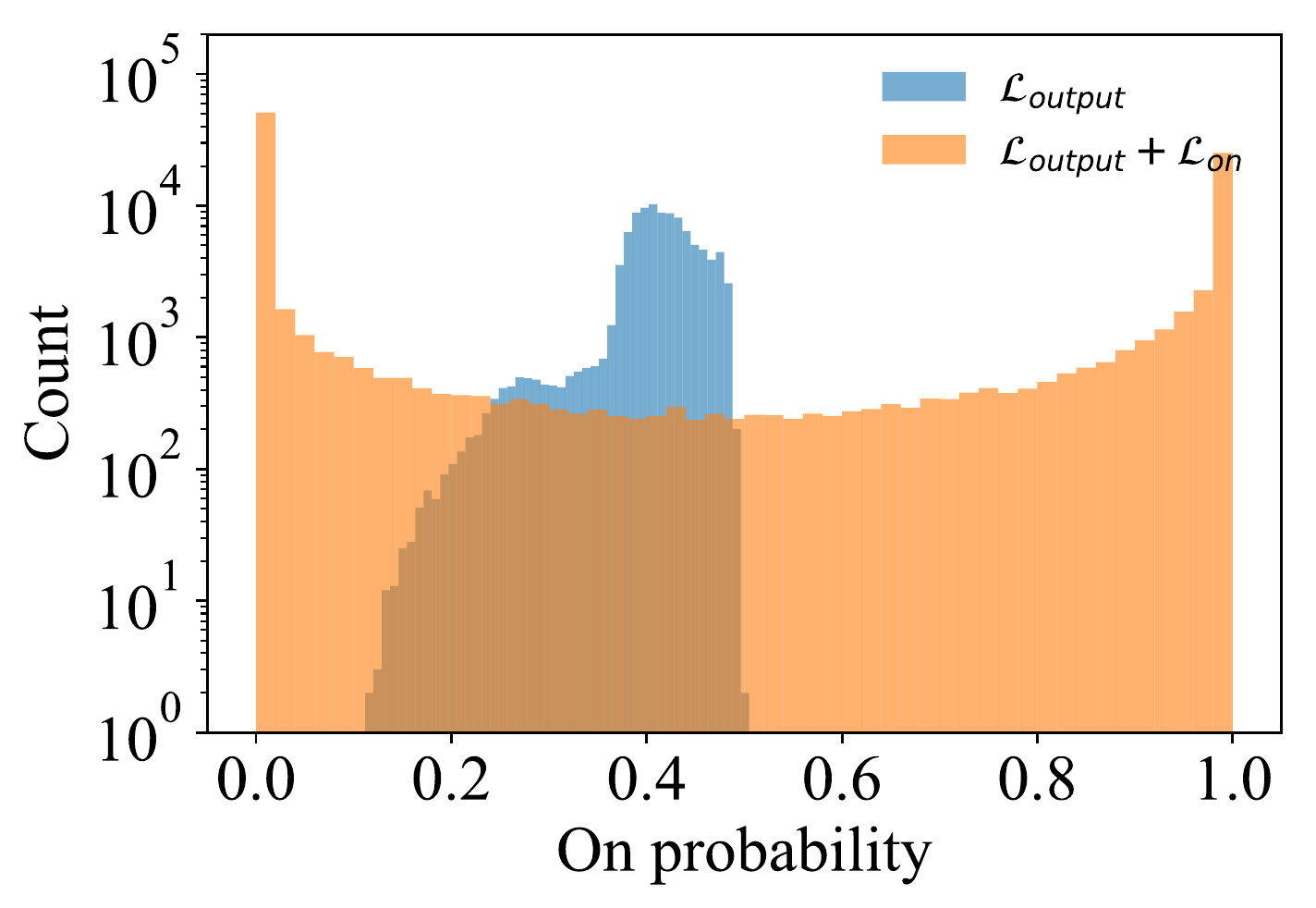}
    \caption*{(b) `On' probability ($\hat{o}_{t}^{i}$) histogram comparison} 
  \end{subfigure}
  \caption{Gating example of UK-DALE fridge. Depending on the choice of loss function, the classification subnetwork's behavior changes. }
  \label{Loutput-comparison}
\end{figure*}

\begin{figure}[h!]
  \centering
    \includegraphics[width=0.40\textwidth]{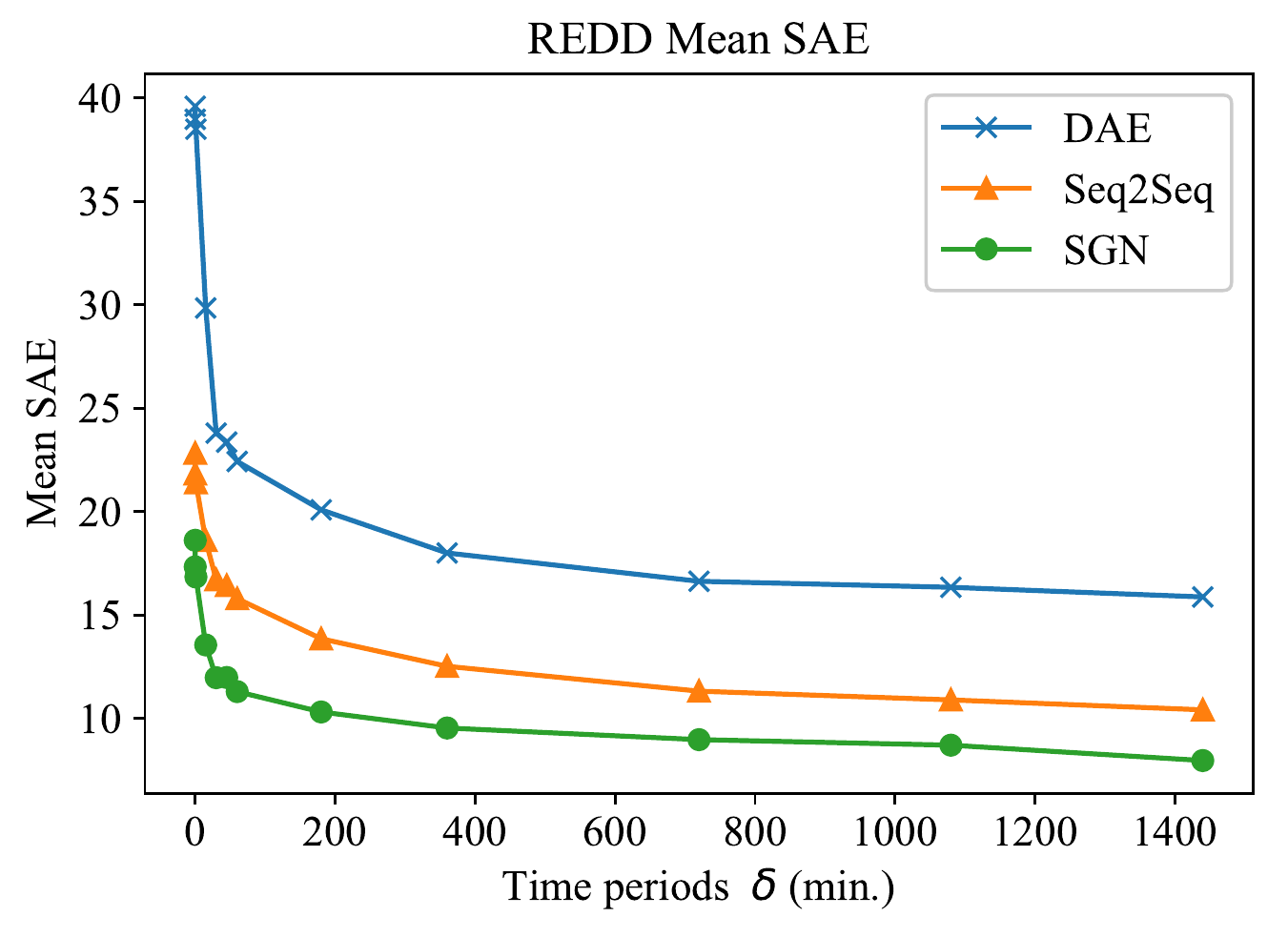}
    \caption{\bm{$SAE_{\delta}$} as a function of \bm{$\delta$} (REDD).}
\label{REDD_SAE}
\end{figure}
 
Figure \ref{gating-examples} shows three examples of how the classification subnetwork interacts with the regression subnetwork. 
In the figure, `SGN before gating' represents $\hat{p}_{t}^{i}$ and `SGN after gating' represents $\hat{p}_{t}^{i} \cdot \hat{o}_{t}^{i}$ respectively.
In the case of the dish washer, gating mainly works for filtering out small noisy regression estimates when the dish washer is really in the `off' state. In the case of the fridge, the regression subnetwork estimates the power consumption of the `on' state even when the fridge is clearly in `off' state. The classification subnetwork is fully responsible for deciding whether it is `on' or `off', and the final output after gating turns out to be quite accurate. 
In the case of the kettle, regression subnetwork seems to have noisy outputs for `off' state just as in the dish washer case. However, the noise is not small valued as in the dish washer. When other appliances are turned on, the kettle's regression subnetwork output seems to be strongly affected and result in a large valued output. However, again the classification subnetwork does its job of gating, and the final output becomes accurate.  

Figure \ref{on-prob} shows the histogram of $\hat{o}_{t}^{i}$, the output of the classification subnetwork. Note that the y-axis is in log scale to properly visualize the large counts around zero and one. Although the main goal is a regression task, a large portion of $\hat{o}_{t}^{i}$ values is concentrated near zero and one. This is because $\mathcal{L}_{on}$ is included in the cost function in addition to $\mathcal{L}_{output}$. Including $\mathcal{L}_{on}$ affects how the regression subnetwork and classification subnetwork cooperate. To illustrate this point, we have generated the same figure as in Figure \ref{gating-examples}(b), but this time without including $\mathcal{L}_{on}$. The results are shown in Figure \ref{Loutput-comparison}. Because $\mathcal{L}_{on}$ is not explicitly included in the cost function, there is no need to perform the classification task well, and the subgated network is free to decide what it wants to do with the two subnetworks. In  Figure \ref{Loutput-comparison}(a), it can be seen that the classification subnetwork became a scaling factor where regression subnetwork simply over-estimates for the `on' states. When the `on' probability histogram is compared as shown in Figure \ref{Loutput-comparison}(b), it is clear that the classification subnetwork has failed to do what it was intended to do. By adding $\mathcal{L}_{on}$ to the cost function, we can prevent the subgated network from being confused and therefore achieve better performance. Going back to Figure \ref{on-prob}, it can be seen that the `on' probability reflects how the appliances really function. For the fridge that cycles between `on' and `off', roughly the same counts can be seen near probability zero and probability one. On the other hand, the counts are skewed for dish washer and kettle that are known to be off most of the time.

\textbf{\bm{$\delta$} sensitivity analysis on \bm{$SAE_{\delta}$}} 
In practice, NILM needs to be used in different ways. Sometimes each appliance's energy usage per hour is needed for the application, but sometimes usage over a day or a month might be needed as well. Depending on the application's requirement, SAE's $\delta$ needs to be chosen appropriately. 
To understand if SGN works well for different choices of $\delta$, we have plotted $SAE_{\delta}$ as a function of $\delta$ in Figure \ref{REDD_SAE}. As expected, the performance becomes better as $\delta$ increases because positive and negative errors can have a chance to cancel out. SGN is the best performing algorithm for the entire range of $\delta$ that was examined.


\section{Conclusions}
In this work, we have proposed subtask gated networks that are effective for the regression problems with on/off states. For the application of non-intrusive load monitoring, performance was evaluated for REDD and UK-DALE datasets. Our proposed solution performed 15-30\% better than the state-of-the-art deep learning solution when MAE and $SAE_{\delta}$ were used as the metrics. The results indicate that combining regression subnetwork and classification subnetwork is a promising direction for  improving the performance of regression tasks that has on/off state inherently.

Future research may be extended in two possible directions. The first is to generalize how to reflect the appliance state information to the model. Our method handles only the on/off states of appliances, but some appliances have multiple states. It might be possible to use the information by creating precise label information beyond on/off states. The second is to extend our method to the data of other domains that have the property similar to on/off. Our method can be applied to any regression problems where we can generate an auxiliary class describing discrete states related to the output.

\section*{Acknowledgments}
This work was supported by the National Research Foundation of Korea (NRF) grant funded by the Korea government (MSIT) (No. NRF-2017R1E1A1A03070560) and by the ICT R\&D program of MSIT/IITP [2016-0-00563]. 



\clearpage
\setcounter{table}{0}
\setcounter{figure}{0}
\subsection{Appendix A. The details of architecture hyperparameters}

SGN in our experiments used two subnetworks, which has the following architecture. Last fully connected layer's activation function exists only in classification subnetwork, which is sigmoid.

\begin{enumerate}[\indent(1)]
        \item Conv layer (filter size = 10, \# of filters = 30, Stride = 1, activation = ReLU)
        \item Conv layer (filter size = 8, \# of filters = 30, Stride = 1, activation = ReLU)
        \item Conv layer (filter size = 6, \# of filters = 40, Stride = 1, activation = ReLU)
        \item Conv layer (filter size = 5, \# of filters = 50, Stride = 1, activation = ReLU)
        \item Conv layer (filter size = 5, \# of filters = 50, Stride = 1, activation = ReLU)
        \item Conv layer (filter size = 5, \# of filters = 50, Stride = 1, activation = ReLU)
        \item Fully connected layer (\# of neurons = 1024, activation = ReLU)
        \item Fully connected layer (\# of neurons = 32, activation = None or Sigmoid)
\end{enumerate}

The details of Kelly's denoising autoencoder are:
\begin{enumerate}[\indent(1)]
        \item Conv layer (filter size = 4, \# of filters = 8, Stride = 1, activation = ReLU)
        \item Fully connected layer (\# of neurons = (sequence length - 3) $\times$ 8, activation = ReLU)
        \item Fully connected layer (\# of neurons = 128, activation = ReLU)
        \item Fully connected layer (\# of neurons = (sequence length - 3) $\times$ 8, activation = ReLU)
        \item Conv layer (filter size = 4, \# of filters = 8, Stride = 1, activation = linear, border mode = valid)
\end{enumerate}

\subsection{Appendix B. Sensitivity to input and output lengths}
Input and output lengths can influence the performance.
To understand the sensitivity, UK-DALE dataset was investigated. 
Recall that deep neural networks can use partial sequences
$\mathbf{x}_{t,s} := (x_t, \cdots, x_{t+s-1})$ and $\mathbf{y}^{i}_{t,s} := (y^{i}_t, \cdots, y^{i}_{t+s-1})$
starting at $t$ with length $s$ as input and output respectively, and that we have included additional windows of length $w$ on both sides of the input to avoid the loss of context information. 

\begin{figure}[t!]
  \centering
    \includegraphics[width=0.48\textwidth]{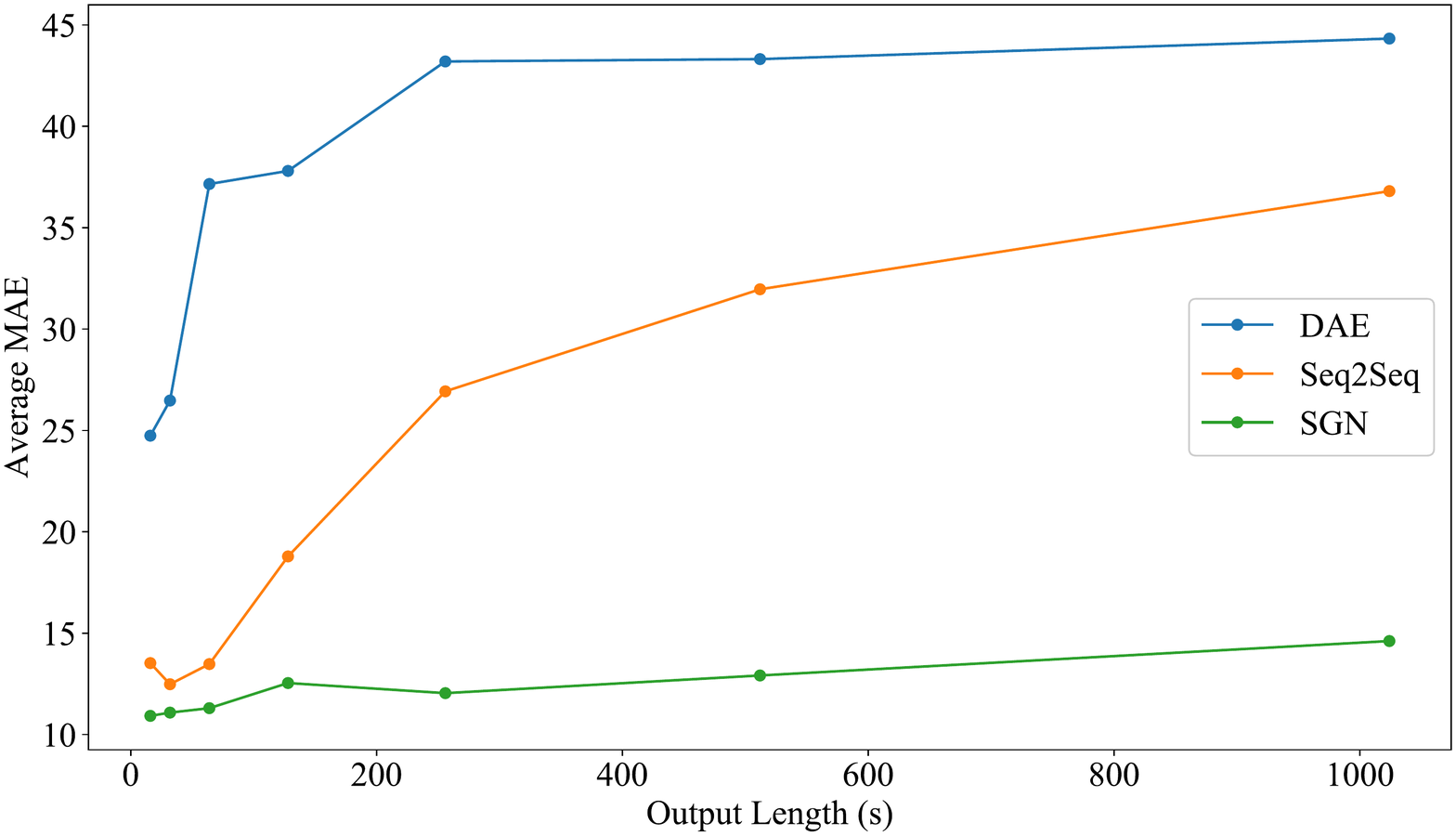}
    \caption{Sensitivity to output $s$ when window length is fixed at $w$ = 200} \label{output_length}
\end{figure}

Figure \ref{output_length} shows the sensitivity result for the output length $s$ when the additional window length is fixed at $w$=200. Note that the input length is equal to $s + 2 w$. The results are shown for the output lengths of \{16, 32, 64, 128, 256, 512, 1024\}.
The figure shows that performance generally becomes worse as the output length becomes comparable or longer than the window length $w$.\\

\begin{figure}[t!]
  \centering
    \includegraphics[width=0.48\textwidth]{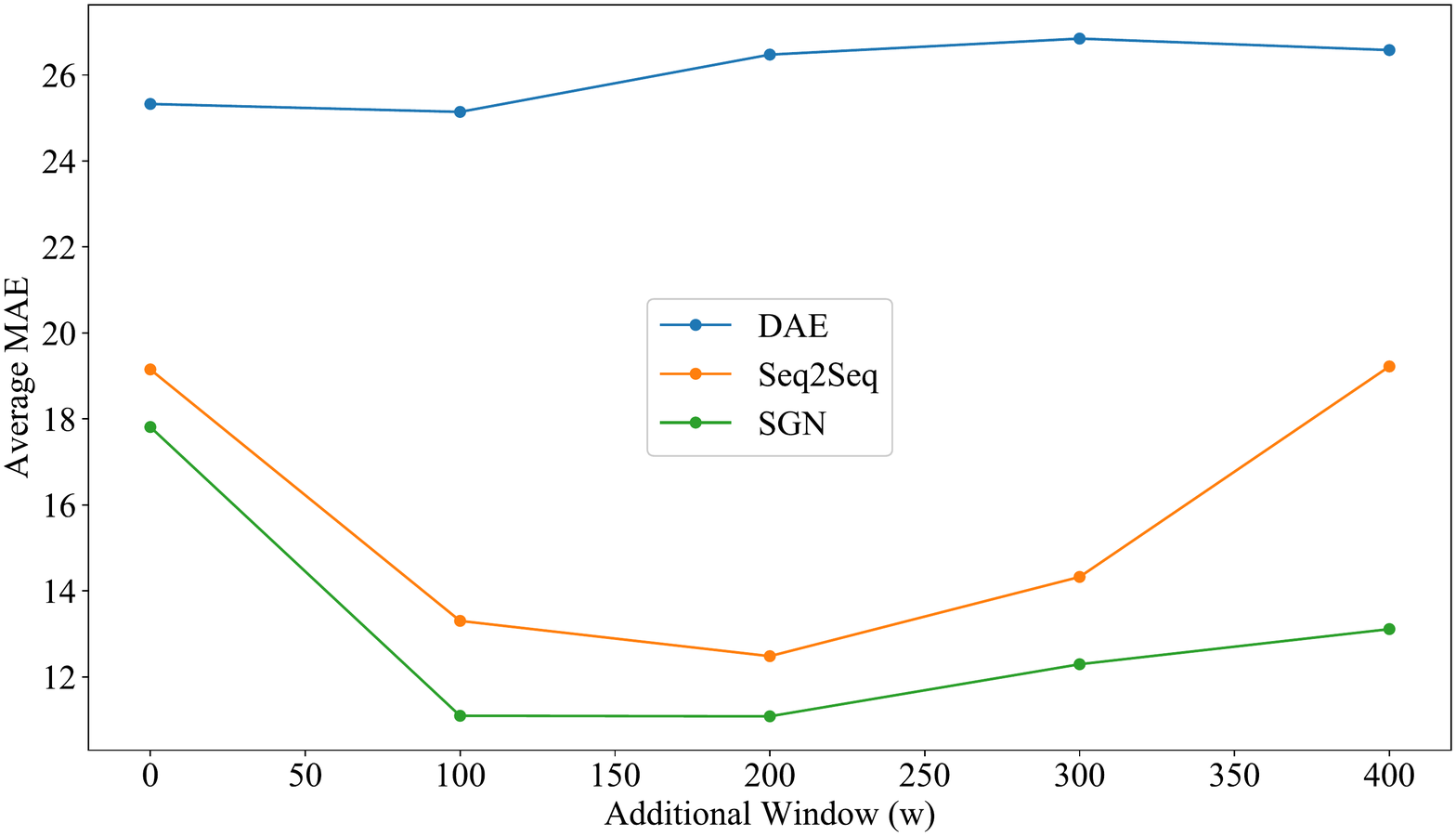}
    \caption{Sensitivity to window length $w$ when output length is fixed at $s = 32$} \label{additional_window}
\end{figure}

Figure \ref{additional_window} shows the sentivity result for the additional window's length $w$. The output length  was fixed at $s=32$. 
In the figure, output lengths of \{0, 100, 200, 300, 400\} were investigated where the corresponding input lengths are \{32, 232, 432, 632, 832\}.
As the window size is increased from 0, the performance is improved for Seq2Seq and SGN. But after reaching $w=200$, the performance starts to degrade with the length of $w$. This indicates that too much extra information hurts the regression performance. 

Overall, SGN outperformed both DAE and Seq2Seq for all the pairs of input and output sizes that were investigated.

\subsection{Appendix C. Training with $\mathcal{L}_{output}$ only}

SGN and its variants can be trained with $ \mathcal {L} _ {output}$ only as mentioned in the main script. For comparison, we experimented with only loss function $\mathcal{L}_ {output} $ in the same setup. Table \ref{Loutput-only-ukdale-table} shows the result. Generally, training with only $\mathcal{L}_{output}$ results in poor performances than training with $\mathcal{L}_{output} + \mathcal{L}_{on}$.

\begin{table}[t!]
  \centering
  \begin{tabular}{ccc}
    \toprule
        Metric & Model & Average \\
    \midrule
    \multirow{4}{*}{MAE}   
                            & SGN ($\mathcal{L}_{output}$ only) & 24.09 \\
                            & SGN-sp  ($\mathcal{L}_{output}$ only) & 24.99 \\    
                            & Hard SGN  ($\mathcal{L}_{output}$ only) & 28.16 \\            
                            & Hard SGN-sp  ($\mathcal{L}_{output}$ only) & 33.55 \\    
    \midrule
    \multirow{4}{*}{$SAE_{\delta}$}  
                            & SGN ($\mathcal{L}_{output}$ only) & 20.16\\
                            & SGN-sp  ($\mathcal{L}_{output}$ only) & 20.05 \\    
                            & Hard SGN  ($\mathcal{L}_{output}$ only) & 22.20 \\            
                            & Hard SGN-sp  ($\mathcal{L}_{output}$ only) & 22.81 \\    
                            
    \bottomrule
  \end{tabular}
  \caption{The results of training with $\mathcal{L}_{output}$ only in REDD.}
  \label{Loutput-only-redd-table}
  \vspace*{.5in}
  \centering
  \begin{tabular}{ccc}
    \toprule
        Metric & Model & Average \\
    \midrule
    \multirow{4}{*}{MAE}    & SGN ($\mathcal{L}_{output}$ only) & 17.17 \\
                            & SGN-sp  ($\mathcal{L}_{output}$ only) & 12.64 \\    
                            & Hard SGN  ($\mathcal{L}_{output}$ only) & 20.12 \\            
                            & Hard SGN-sp  ($\mathcal{L}_{output}$ only) & 30.25 \\    
                            
    \midrule
    \multirow{4}{*}{$SAE_{\delta}$} 
                            & SGN ($\mathcal{L}_{output}$ only) & 14.12 \\
                            & SGN-sp  ($\mathcal{L}_{output}$ only) & 8.81 \\    
                            & Hard SGN  ($\mathcal{L}_{output}$ only) & 17.00 \\            
                            & Hard SGN-sp  ($\mathcal{L}_{output}$ only) & 23.38 \\    
    \bottomrule
  \end{tabular}
  \caption{The results of training with $\mathcal{L}_{output}$ only in UK-DALE.}
  \label{Loutput-only-ukdale-table}
  \vspace*{7in}
\end{table}

\end{document}